\pdfoutput=1

\documentclass[11pt]{article}

\usepackage[final]{acl}

\usepackage{times}
\usepackage{latexsym}
\usepackage{amsmath}

\usepackage[T1]{fontenc}
\usepackage{tablefootnote}
\usepackage{threeparttable}

\usepackage[utf8]{inputenc}

\usepackage{microtype}

\usepackage{inconsolata}

\usepackage{graphicx}

\usepackage{cleveref}
\usepackage{multicol}
\usepackage{multirow}
\usepackage{wrapfig}
\usepackage{xspace}
\usepackage{geometry}
\usepackage{booktabs} 
\usepackage{array} 
\usepackage{float}

\NewDocumentCommand{\var}{O{s} m O{}}{%
  \ensuremath{#1_{#2}^{#3}}
}
\usepackage{siunitx}

\newcommand{\commentout}[1]{}

\definecolor{light-gray}{gray}{0.80}

\newcommand\fref{Fig.~\ref}

\newcommand\sref{\S~\ref}

\usepackage{amsthm}

\usepackage{amsthm}

\newtheorem{mytheorem}{Theorem}
\newtheorem{assumption}[mytheorem]{Assumption}
\newtheorem{lemma}[mytheorem]{Lemma}
\newtheorem{remk}[mytheorem]{Remark}

\newcommand{\name}{MedCite\xspace}

\title{MedCite: Can Language Models Generate Verifiable Text for Medicine?}

\author{
  \textbf{Xiao Wang}\textsuperscript{1},
  \textbf{Mengjue Tan}\textsuperscript{2},
  \textbf{Qiao Jin}\textsuperscript{3},
  \textbf{Guangzhi Xiong}\textsuperscript{4},
  \textbf{Yu Hu}\textsuperscript{5},\\
  \textbf{Aidong Zhang}\textsuperscript{4},
  \textbf{Zhiyong Lu}\textsuperscript{3},
  \textbf{Minjia Zhang}\textsuperscript{1}
  \\
  \textsuperscript{1}SSAIL Lab, University of Illinois at Urbana-Champaign \quad
  \textsuperscript{2}Brown University \\
  \textsuperscript{3}National Library of Medicine, NIH \quad
  \textsuperscript{4}University of Virginia \quad
  \textsuperscript{5}Microsoft \\
  \texttt{\{xiaow4, minjiaz\}@illinois.edu}\\
  \texttt{mengjue\_tan@brown.edu} \quad
  \texttt{\{qiao.jin, zhiyong.lu\}@nih.gov}\\
  \texttt{\{hhu4zu, aidong\}@virginia.edu} \quad
  \texttt{yuhu@microsoft.com}
}
\begin{document}
\maketitle

\begin{abstract}

Existing LLM-based medical question-answering systems lack citation generation and evaluation capabilities, raising concerns about their adoption in practice. In this work, we introduce \name, the first end-to-end framework that facilitates the design and evaluation of citation generation with LLMs for medical tasks. Meanwhile, we introduce a novel multi-pass retrieval-citation method that generates high-quality citations.   
Our evaluation highlights the challenges and opportunities of citation generation for medical tasks, while identifying important design choices that have a significant impact on the final citation quality. Our proposed method achieves superior citation precision and recall improvements compared to strong baseline methods, and we show that evaluation results correlate well with annotation results from professional experts. 

\end{abstract}

\section{Introduction}
\label{sec:intro}

Large Language Models (LLMs) have demonstrated remarkable capabilities in various natural language processing tasks, such as question answering (QA) and instruction following~\citep{scaleing-law-nlp,emergent-capability,cot}. Progress in LLMs has also enabled the development of medical agents that understand language used by patients and physicians, offering rich just-in-time assistance~\cite{llm-encode-clinical-knowledge,med-palm,temsah2023chatgpt,tangadulrat2023using,maples2024loneliness}. 

\begin{figure}[t]
    \centering
    \includegraphics[width=\columnwidth]{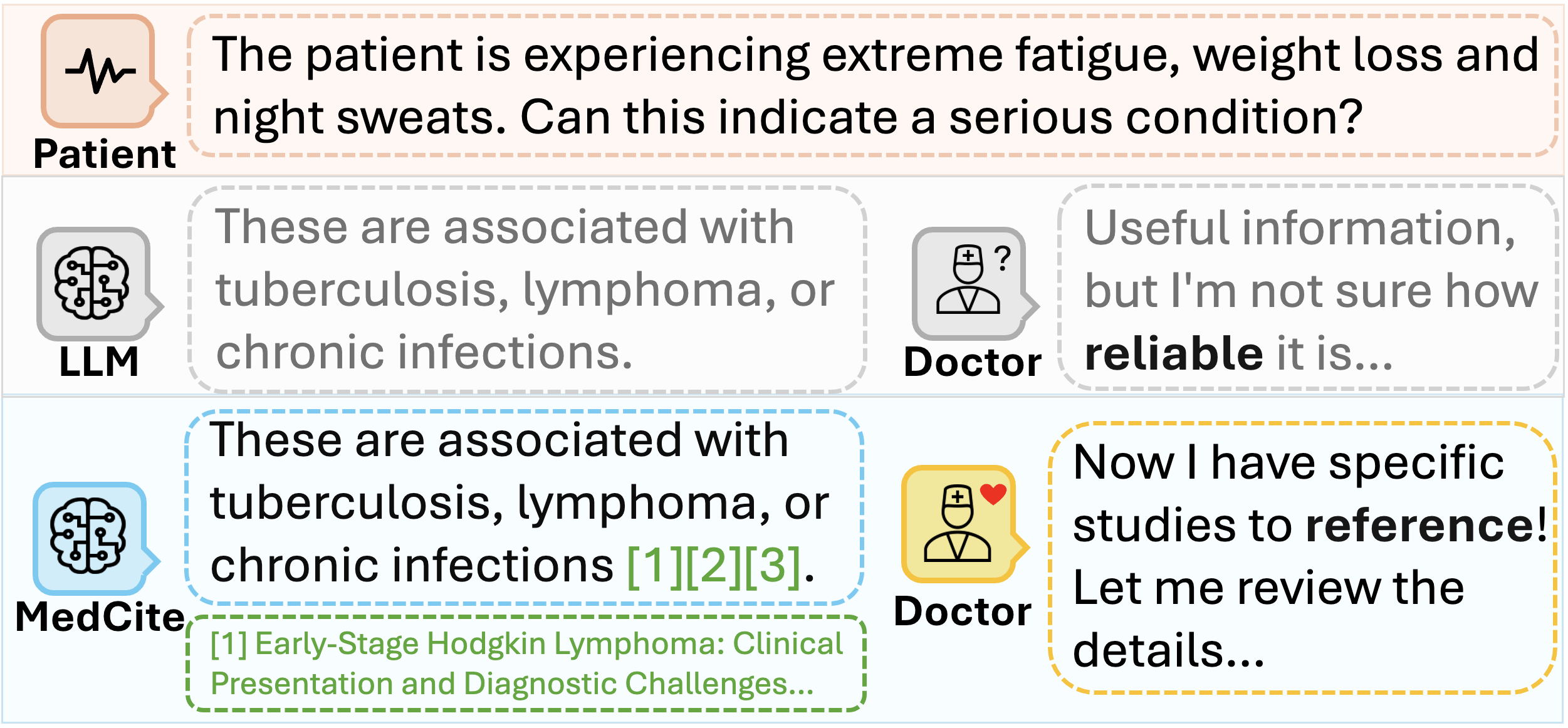}
    \caption{Medical QA system comparison. State-of-the-art systems generate answers without citations. \name not only generates answers but also associates each answer with citations, improving the verifiability and trustworthiness of the medical system.}
    \label{fig:1}
\end{figure}

While the early signs are positive, current LLM-powered medical QA systems still have multiple limitations. For example, medical data often contains sensitive information, such as personal health records, which require strict ethical compliance for training LLMs~\cite{llm-medical-approval}. Furthermore, trustworthiness is particularly important in the medical field. Issues such as hallucination, where the model generates information that is incorrect or misleading, pose significant challenges to the reliability of LLM-based medical systems~\cite{pal2023med,ahmad2023creating,llm-trustworthy}. 
To overcome the issue, researchers and practitioners have studied retrieval-augmented generation (RAG)~\citep{xiong2024benchmarking,llm-bioqa}, which combines LLMs with information retrieval from external trustworthy data sources~\citep{canese2013pubmed}. By providing the model with accurate and relevant medical knowledge, these systems allow LLMs to maintain relevance in responses. 

\begin{figure}[!ht]
    \centering
    \includegraphics[width=0.45\textwidth]{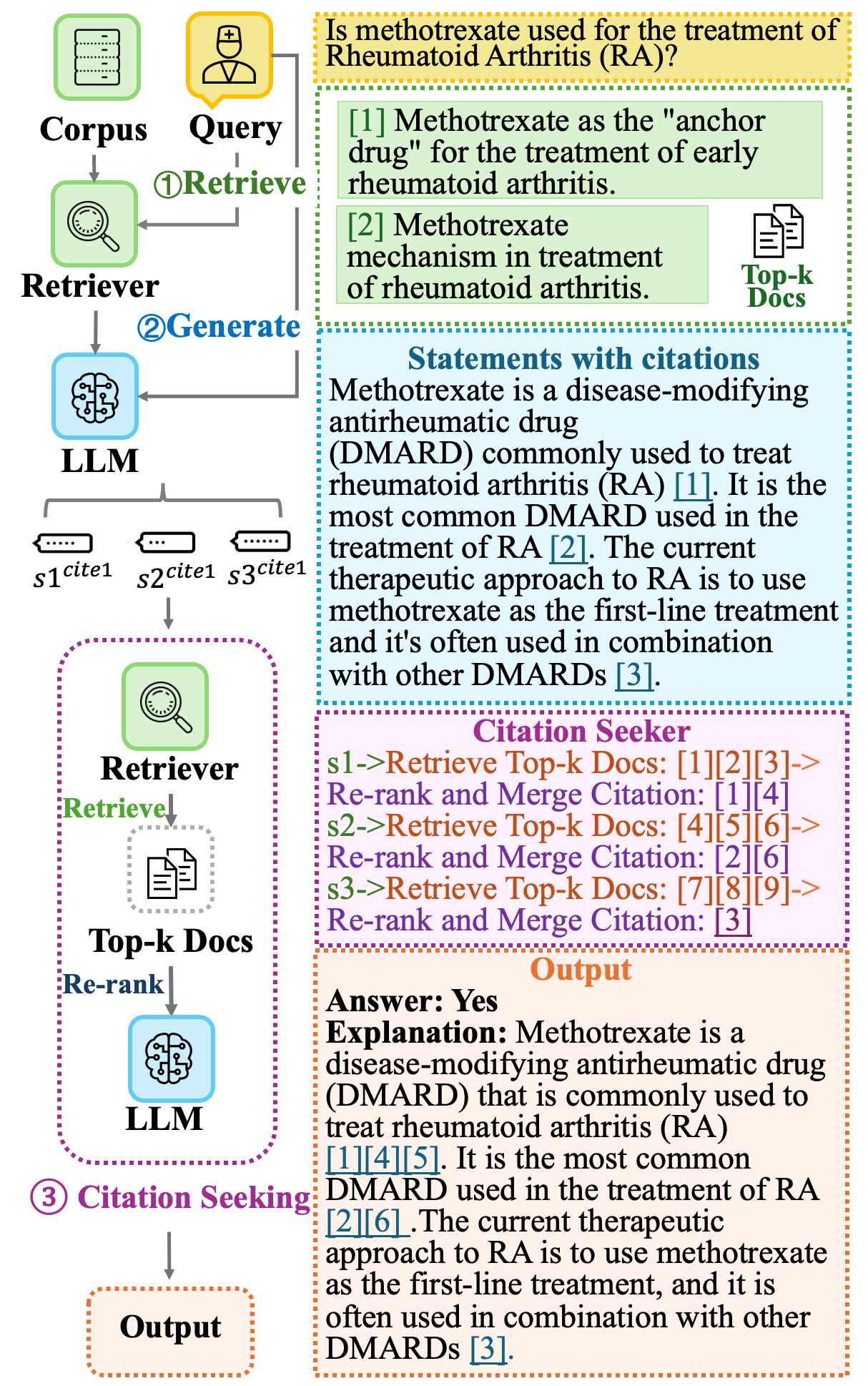}
    \caption{The overview figure of \name.}
    \label{fig:overview}
\end{figure}

Despite promising results, existing methods lack \emph{verifiability}~\cite{verifiability}, meaning that the answers provided are not backed by reliable sources or evidence. This can lead to misinformation and potentially harmful consequences if incorrect medical advice is followed. For instance, as shown in \fref{fig:1}, when providing a diagnosis based on a list of symptoms without any references, the accuracy of prognosis or treatment recommendations cannot be assured, which creates a sense of uncertainty, leading to suboptimal or even harmful decisions.

One promising approach to mitigate the verifiability concern is through \emph{attribution}~\cite{attributed-qa,citation-llm}, i.e., associating statements with \emph{citations}, which offers the system more credibility and accountability while providing users a way to explore the source in greater depth and verify the information source.
However, although there are prior efforts that analyze citation capabilities through LLMs on general domain QA tasks~\cite{verifiability,alce,attribution-ir-evaluation}, citing sentences for medicine is especially challenging and not widely adopted in practice due to the following reasons.

First, existing works on medical QA often leverage multi-choice accuracy to benchmark and evaluate their performance~\cite{xiong2024benchmarking,llm-bioqa,enhance-healthcare}, which focuses on evaluating the ability to select the correct answer from a set of given options. 
However, citation generation is more challenging due to its open-ended nature. For example, when prescribing medication for a rare genetic disorder or planning surgery for a patient, both physicians and patients have to rely on richer information. Therefore, a query can have multiple answers, supported by multiple possible sources. These aspects are important to consider for the evaluation of citation generation methods, but existing medicine QA frameworks do not inherently account for them. 

Second, there is a huge design space for citation generation with complex interactions among retrievers~\cite{self-rag,dense-retriever}, backbone LLMs~\cite{llama3,ultra-medical,gpt4}, and citation generation algorithms~\cite{alce}. Therefore, identifying key contributing factors is challenging. However, the analysis is crucial for developing strategies to improve the verifiability of medical systems.

Third, while contributions continue to rise in this field~\cite{alce,xiong2024benchmarking,llm-bioqa,enhance-healthcare}, there is a noticeable deficiency in open-source frameworks useful for designing, developing, and evaluating citation generation quality for medical tasks. The existing citation evaluation frameworks are constructed from generic questions, where the selection of metrics and evaluators for medical tasks remains an open question. Moreover, it is quite costly to obtain high-quality medical expert annotations,  
which demands a high-quality classifier to judge whether a citation attributes to a statement. 

In this paper, we tackle these challenges with the hope of fostering research in improving verifiability for medicine systems. 
In particular, our contributions are as follows:

\begin{itemize}
    \item An in-depth study of 
    different citation methods and key design components for medical tasks using LLMs, ranging from text generation methods, information retrieval methods, to citation attribution methods. Our study disentangles the importance of different factors from the backbone LLM.   
    \item We present \name, the first end-to-end system for enabling LLMs to generate verifiable texts for medical QA systems with automatic evaluation. Meanwhile, we introduce a novel multi-pass retrieval-citation method that conciliates retrieval-augmented generation and post-generation citation.
    \item A comprehensive evaluation of \name across different LLMs, which shows that \name outperforms existing methods in both text generation and citation generation quality by up to 47.39\% recall and 31.61\% precision respectively. We conduct human evaluation by having medical doctors verify the attribution results. The results show that our automatic evaluation pipeline correlates well with domain expert judgments, demonstrating the effectiveness of efficient and automatic citation evaluation for medicine. 
\end{itemize}

Our code are available at \url{https://supercomputing-system-ai-lab.github.io/projects/medcite/}

\section{Related Work}
\label{sec:related_short}

\paragraph{Biomedical Question Answering.}
Biomedical QA aims to answer clinical or biomedical questions using NLP techniques. Early systems were rule-based~\cite{medqa,askhermes}, relying on structured ontologies but lacking scalability. Later, domain-specific language models such as BioBERT~\cite{biobert} and ClinicalBERT~\cite{clinical-bert}, built on BERT~\cite{bert}, brought performance gains across biomedical QA benchmarks~\cite{llm-bioqa}. Recently, generative models like GPT-3.5/4~\cite{gpt-3,gpt4} and Med-Gemini~\cite{med-gemini} enabled flexible answer generation without predefined options. However, these models can hallucinate, motivating the adoption of retrieval-augmented generation (RAG) to ground outputs on retrieved documents~\cite{lozano2023clinfo,xiong2024benchmarking,llm-bioqa,enhance-healthcare,zakka2024almanac,i-medrag}. Our work focuses specifically on improving the verifiability of generated biomedical responses.

\paragraph{Citation Methods for LLM Generation.}
Adding citations to LLM-generated text has become an active area of research. Although LLMs such as ChatGPT~\cite{gpt-3,lamda,gemini-google,gpt4,gpt4o} can be prompted to produce citations, these are often inaccurate or fabricated~\cite{chatgpt-attribution-hallucination}. Direct model-driven attribution~\cite{recitation-augmented,agrawal2023language,according-to} enables LLMs to cite autonomously but lacks reliability. Retrieval-based methods improve grounding: post-retrieval generation (PRG) retrieves relevant documents before generation~\cite{rag-pretraining,retrieval-trillions-tokens,reddy2023smartbook}, while post-generation citation (PGC) retrieves evidence after answer generation~\cite{retrieve-supporting-evidence}. Though more robust, both approaches add complexity~\cite{gao2023rarr}, and may still fall short on biomedical tasks requiring precise attribution. Our hybrid double-pass citation method integrates RAG with post-generation refinement to address these limitations. Another direction involves fine-tuning LLMs on curated or synthetic citation data to improve citation quality~\cite{ye2024effectivelargelanguagemodel}. Lastly, evaluation protocols for citation accuracy are being developed~\cite{measuring-nlg-attribution,alce,towards-verifiable-generation}, though existing work largely focuses on general-domain content. In contrast, our evaluation is medicine-specific and considers domain-aware strategies for retrieval and citation generation.

\paragraph{Evaluation Frameworks in the Medical Domain.}
\citet{wu2024llmsciterelevantmedical} proposed a citation evaluation pipeline based on URL-retrieval and showed that even top LLMs like GPT-4 (RAG) often failed to provide fully supported answers. While both our work and theirs aim to improve citation reliability, they focus on prompting API-based models for online URLs, whereas we propose a modular citation system that improves biomedical grounding. As detailed in Section~\ref{sec:method}, we show why parametric citation methods are unsuitable for open-source LLMs due to issues like hallucinated URLs and lack of access to verifiable sources. Our method leverages hierarchical retrieval from trusted medical corpora such as PubMed and incorporates multi-pass citation to ensure attribution of specialized content, including drug names and genomic markers.
A more comprehensive discussion of related work is provided in Appendix~\ref{sec:related}.

\section{Problem Setup}
\label{sec:task_setup}

In this section, we first formulate the citation generation task for biomedical QA and then give an overview of the approaches that we will examine experimentally in the following section.

\subsection{Problem Objective}

The objective is to develop a system that automatically adds relevant and accurate citations to text statements generated by a large language model. In particular, the inputs to the system include a user query $q$, an LLM $\Phi$, an external database $D$, which contains ground truth documents. The outputs of the system include a generated text passage, which contains a list of statements $S$ = $\{s_0, s_1,...,s_n\}$ by $\Phi$. For each statement $s_i$, a set of in-line citations $C_i$ = $\{c_i^0, c_i^1,..., \}$, where $c_i^j \in $D, is assigned to it. 

\subsection{Dataset}

Following prior work~\cite{bolton2024biomedlm27bparameterlanguage,yasunaga2022deepbidirectionallanguageknowledgegraph,xiong2024benchmarking}, we use the BioASQ-Y/N dataset~\cite{bioasq}, which is a commonly used dataset for benchmarking biomedical question answering systems. The dataset consists of questions, human-annotated answers, and relevant contexts that provide the necessary information to answer the questions. The BioASQ-Y/N dataset has three characteristics that motivate us to use it for the study: 1) Unlike other datasets used for medical QA~\cite{medqa-us,mmlu-med,medmcqa}, which are primarily multi-choice QA tasks, BioASQ-Y/N not only provides option choices (Yes/No) but also a gold set of answers w.r.t the informativeness of answer statements. 2) BioASQ-Y/N provides ground truth labels of the supporting documents for each question. Meanwhile, it can be easily modified to answer questions without the ground-truth documents provided, which represents a more realistic medical setting. 3) It has not so far been used by existing generic citation methods. Apart from BioASQ used for the analysis in Section~\ref{sec:analysis}, we 
also include PubMedQA~\cite{pubmedqa} in Section~\ref{sec:res}. We include the details of the datasets and hyperparameters in Appendix~\ref{appendix:extra-exp-details}. 
On the external database side, we primarily consider PubMed database~\cite{canese2013pubmed}, which contains 24.6 million biomedical documents
vetted by medical professionals. This vast database provides access to a wealthy source of precise and legitimate documents LLM-generated text can attribute to. 

\subsection{Evaluation Metrics}
\label{subsec:metrics}

For medicine QA, evaluating both text and citation generation quality is crucial to ensure that the outputs of LLMs are not only coherent and relevant but also well-supported by accurate citations. As such, we consider the following aspects. 
    
\paragraph{Answer correctness.} Different from multi-choice QA, real medical systems often generate long and open-ended answers. Therefore, we use
ROUGE-L~\cite{lin2004rouge} and MAUVE~\cite{pillutla2021mauve} to evaluate the correctness and relevance of the answer based on the ground truth answer. We still let the model generate a Yes/No answer in addition to the long answer, such that we can make comparisons with existing non-citation methods. 

\paragraph{Citation quality.} We evaluate citation quality at the statement level, which offers a finer-grained view than question-level document retrieval. This better captures whether each generated statement is grounded in medical evidence.

We consider an attribution judge $\emph{Attr}: \mathcal{X}, \mathcal{Y} \rightarrow \{0, 0.5, 1\}$ that outputs 1 if the statements $\mathcal{X}$ can be fully attributed to the statements $\mathcal{Y}$, i.e., $\mathcal{Y}$ is the source of $\mathcal{X}$, 0.5 if $\mathcal{X}$ can be partially attributed to $\mathcal{Y}$, and 0 otherwise. To justify the introduction of partial support, we refer to findings from recent studies, such as \citet{Whrl2024WhatMM}, which showed that in medical fact-checking tasks, 62.4\% of claims were partially supported by evidence. This highlights the importance of capturing partial attribution, as it is a frequent occurrence in real-world medical statements. 

For the use of citations in medical QA, an answer can have multiple verifiable statements, and multiple citations may be attached to support one statement. S
With the attribution judge, we measure citation qualities with two metrics: 
\emph{citation recall}, 
and \emph{citation precision}.
Both citation recall and precision heavily affect the usability of medical QA, as a high recall means that the generated responses are well supported by evidence, and a high precision indicates that the assigned citations have high quality that can be used to verify the truthfulness of the generated texts. 
For simplicity, let us consider a single statement \( s \) with \( n \) citations \( c_1, c_2, \dots, c_n \), where each of them is a set of axioms.

\paragraph{Citation recall.} We define recall as a statement-level metric, which measures whether all the information in the statement is fully supported by the citations. 
For recall = 1, every axiom (fact) in \( A_s \) must be present in the collection of axioms provided by all the citations. 

In our experiments, we use the concatenation of the citation documents to represent the union of citations and make a judgment on whether the statement can or cannot be fully supported by the concatenated citations. Using the attribution judge defined above, we have \(\text{Recall}(s,c_1,\cdots,c_n) = 1\) if and only if \(\emph{Attr}(s, \bigcup_{i=1}^n c_i) = 1\).
We then average over all statements to get the final recall score of an answer passage.

\paragraph{Citation precision.} Following previous research \cite{verifiability}, we define the precision metric as a citation-level measurement, which assesses if each individual citation contributes to supporting the statement. Precision is 1 for a citation if it either fully or partially supports the statement by containing at least some of the necessary axioms from the statement. 
The precision of \(c_i\) for \(s\) is computed as 1 if and only if \(\emph{Attr}(s, c_i) > 0\). When having multiple statements, we compute its citation precision by averaging the precision scores of all citations in it.

\paragraph{Citation F$_1$.} We use citation F$_1$~\cite{verifiability} to measure the combined citation precision and recall via: $F_1=2\times\frac{citation\ precision \times citation\ recall}{citation\ precision + citation\ recall}$.



\section{Citation Procedure Analysis}
\label{sec:analysis}

This section explores and quantifies which choices are important for successfully citing sentences for medical tasks. Given that each component can be varied, we investigate how each of these components impacts the citation generation quality while isolating the other components. Unless otherwise specified, we use Llama-3-8B-I. for the experiments in this section.
\subsection{Parametric vs. Non-Parametric Citation}
\label{subsec:parametric-citation}

Recent LLMs can be prompted to include citations in the text they generate by relying on its parametric contents, i.e., information internalized from the training data. Given this advancement, one question naturally arises: \emph{can we rely on LLMs to self-cite their generated sentences?}
We compare this strategy with non-parametric citation where we generate citations by purely relying on non-parametric information-retrieval (IR) contents, e.g., PubMed. In particular, for parametric citation, we generate a prompt that includes the user question, and a directive instruction for LLM to generate answers while adding in-line citations in formatted output for each statement. 
In this case, the model solely depends on its pre-training data to generate citations. For the non-parametric citation, we let LLM to directly generate an answer without citations. Then we use a dense retriever MedCPT~\cite{medcpt} to retrieve a list of relevant document (e.g., top-3) from $D$ for each answer statement and those documents as in-line citations. The prompts used can be found in Appendix~\ref{subsec:prompts}. 

\begin{table}[!ht]
\tiny
\centering
\newcommand{\smallcolspc}{\hspace*{0.24em}}
\newcommand{\colspc}{\hspace*{0.24em}}
\begin{tabular}{@{\smallcolspc}c@{\smallcolspc}c@{\smallcolspc}c@{\smallcolspc}c@{\smallcolspc}c@{\smallcolspc}c@{\smallcolspc}c@{\smallcolspc}}
\toprule
\multirow{2}{1cm}{\textbf{Citation Method}} & \multirow{2}{*}{\textbf{Model}} & \multirow{2}{1cm}{\textbf{Accuracy (EM)}} & \multicolumn{2}{c}{\textbf{Text Quality}} & \multicolumn{2}{c}{\textbf{Citation Quality}} \\
\cmidrule(lr){4-5} \cmidrule(lr){6-7}
                         &                        &                        & \textbf{MAUVE} & \textbf{ROUGE-L} & \textbf{Rec.} & \textbf{Prec.} \\ 
\midrule
\multirow{3}{*}{Parametric (LLM)}    & Llama-3-8B-I.  & 74.76 & 61.94 & 17.72 & / & / \\ 
                               & UltraMedical & 69.09 & 67.70 & 13.96 & / & / \\ 
                               & GPT-4o & \textbf{88.51} & \textbf{74.82} & \textbf{20.03} & / & / \\ 
\midrule
\multirow{3}{*}{Non-parametric (IR)} & Llama-3-8B-I. & 73.95 & 65.31 & 19.05 & 60.89 & 53.90 \\ 
                                & UltraMedical & 68.12 & 51.18 & 12.69 & 52.48 & 62.32 \\ 
                                & GPT-4o & \textbf{87.70} & \textbf{70.15} & \textbf{20.20} & \textbf{79.72} & \textbf{80.95} \\ 
\bottomrule
\end{tabular}%
\caption{Comparison of parametric (LLM) vs. non-parametric (IR) citation methods across different LLMs.}
\label{tbl:parametric-vs-nonparametric}
\end{table}

\begin{figure*}[h]
    \centering
    \begin{minipage}[t]{0.30\textwidth}
        \centering
        \includegraphics[width=\textwidth,height=\textheight,keepaspectratio]{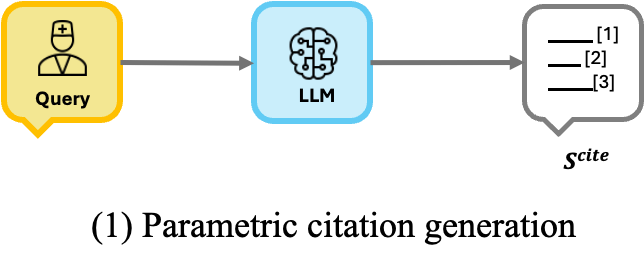}
    \end{minipage}
    \hfill
    \begin{minipage}[t]{0.60\textwidth}
        \centering
        \includegraphics[width=\textwidth,height=\textheight,keepaspectratio]{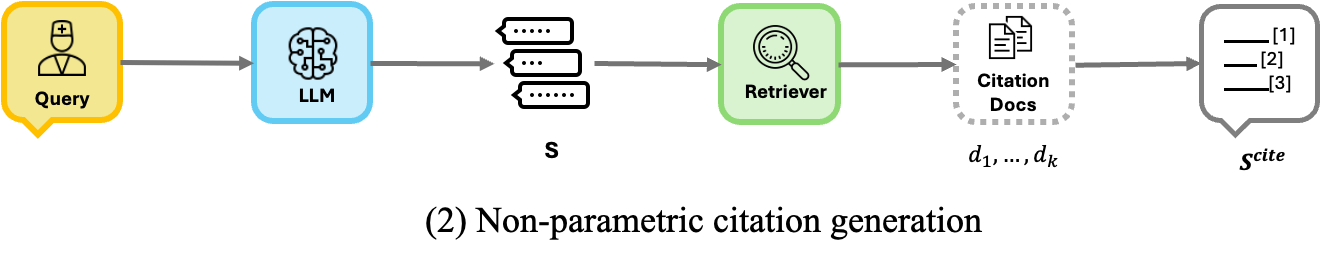}
    \end{minipage}
    
    \caption{Comparison of parametric (LLM) and non-parametric (IR) citation generation pipelines.}
    \label{fig:prg}
\end{figure*}

Table~\ref{tbl:parametric-vs-nonparametric} compares the parametric vs. non-parametric citation results across different LLMs. We find that while LLMs have made significant strides in understanding and following human instructions, they do have limitations when it comes to generate citations in medical settings.  In particular, both Llama-3-8B-I. and UltraMedical cannot follow those instructions accurately. As a result, the generated citations are either incorrect, fabricated, or ill-formatted, and even though a small proportion of them do exist, such citations might not be freely accessible (e.g., some scientific articles are behind a paywall). As such, without API access to the content of any scientific articles generated as citations by parametric methods, it is challenging to automatically evaluate their quality. This is unsurprising because these models are still trained on next-token prediction, and LLM needs to extrapolate the citation information with its pretraining knowledge or hallucination.
Interestingly, GPT-4o not only achieves the highest accuracy on the BioASQ task, but it can also consistently follow the instructions to generate well-formatted citations. However, the references GPT-4o generated are outdated (all before year 2018), making it hard to include new studies. We include several examples of generated citations in Appendix~\ref{subsec:ill-formated-citations}.
This observation highlights a critical limitation of the parametric-only approach when applied to citation generation, particularly in the medicine domain with public LLMs. Given these limitations, we focus on non-parametric citation methods using trusted datasets like PubMed in the remainder of the experiments.

\subsection{RAG Makes Better Citations}
\label{subsec:rag-analysis}

While non-parametric citation improves the citation quality, the answer statement generation still relies on the pre-training data itself. Therefore, the answers can be based on outdated or incomplete medical data. Despite multiple recent papers observing that adding retrieval-augmented generation (RAG) helps improve LLM to better understand biomedical tasks, producing higher accuracy than non-RAG based approaches~\cite{xiong2024benchmarking,llm-bioqa,enhance-healthcare}, experiments on how RAG affects both text and citation quality are rarely reported.  
We dig deeper into the role of RAG in citation generation by comparing several different methods. For all methods, we use the same dense MedCPT based retriever to assign top-k (e.g., top-3) relevant documents as citations.

\noindent \textbf{Non-RAG (CoT)}: We perform chain-of-thought (CoT) prompting~\cite{cot} to leverage the reasoning capability of LLMs to provide an answer (e.g., a polar Yes/No answer) and text explanations $S$ to the question $q$. This is similar as the method in \cite{xiong2024benchmarking}, but no supporting context is retrieved from the external database. 

\noindent \textbf{RAG}: We first retrieve a shortlist of top-$k$ supporting documents $\{d_1,...,d_k\}$ to the query $q$ from $D$. We then feed the concatenate shortlist documents together with $q$ to the LLM, and instruct the LLM to generate the answer and text explanations $S$. 

\noindent \textbf{RAG w. Oracle}: Similar to the above configuration, but using the ground truth supporting documents (i.e., assuming a perfect retriever) in BioASQ for each question. 

\begin{table}[!ht]
\centering
\newcommand{\smallcolspc}{\hspace*{0.32em}} 
\newcommand{\colspc}{\hspace*{0.32em}}
\scriptsize
\begin{tabular}{@{\smallcolspc}l@{\smallcolspc}c@{\smallcolspc}c@{\smallcolspc}c@{\smallcolspc}c@{\smallcolspc}c@{\smallcolspc}}
\toprule
\multirow{2}{1cm}{\textbf{Retrieval Method}} & \multirow{2}{1cm}{\textbf{Accuracy (EM)}} & \multicolumn{2}{c}{\textbf{Text Quality}} & \multicolumn{2}{c}{\textbf{Citation Quality}} \\
\cmidrule(lr){3-4} \cmidrule(lr){5-6}
                        &                        & \textbf{MAUVE} & \textbf{ROUGE-L} & \textbf{Recall} & \textbf{Precision} \\ 
\midrule
Non-RAG                 & 71.36                   & 53.24 & 18.07 & \textbf{59.05}          & \textbf{52.93 }          \\ 
\midrule
RAG                     & 82.85                   & 52.22 & 14.79 & 49.01          & 42.77           \\ 
\midrule
RAG w. Oracle           & \textbf{94.34}                   & \textbf{63.45} & \textbf{20.63} & 57.46          & 43.20           \\ 
\bottomrule
\end{tabular}
\caption{Comparison of RAG and Non-RAG methods for citation generation.}
\label{tbl:rag-method-comparison}
\end{table}

Table~\ref{tbl:rag-method-comparison} shows a comparison of non-RAG and RAG methods for medicine. Interestingly, we observe that without RAG, the correctness of generated answers tend to be low (71.36). However, the citation recall and precision are relatively high. Conversely, integrating RAG leads to a significant increase in answer correctness while resulting a decrease in citation recall and precision. This is because the citation quality metrics only assess whether the LLMs' generated statements are \emph{supported} by verifiable sources, rather than directly assessing the \emph{correctness} of each statement. 
Therefore, it is possible that the identified citations can still support a hallucinated statement, even though that statement is irrelevant to the user's question. This finding suggests that we need to holistically assess LLM's capability for both text and citation generation. Specifically, we treat the correctness of the answers (e.g., accuracy and text quality) as a prerequisite of the evaluation of citations, and enabling citation capabilities should not compromise the quality of answer generation.

Table~\ref{tbl:rag-method-comparison} also shows that by using the ground truth documents (oracle), the best obtainable results for accuracy using RAG can go up to 94.34\% and the citation recall and precision can go up to 57.46\% and 43.20\% respectively, leaving room to investigate better retrieval augmentation methods. 
Nevertheless, these results indicate that RAG is crucial to generating context relevant texts and is a critical step for getting high-quality citations. Therefore, we use RAG in the remainder of our experiments.

\subsection{The Efficacy of Citation Seeker}
\label{subsec:citation-seeking}

Till now we have fixed the choice of the citation seeker, i.e., how to find relevant documents and assign them as citations to a statement. However, one may wonder how the choice of citation-seeking methods affects citation quality. To investigate this, we consider the following strategies:

\noindent \textbf{Pre-generation shortlist + LLM rerank}. For each generated statement, we instruct LLM to assign a document retrieved as one of the supporting documents from the pre-generation retrieval. No additional retrieval is needed in this case.

\noindent \textbf{Retriever-only re-retrieval}. For each generated statement, we relaunch the retriever to retrieve top-k documents relevant to the statement from $D$ and append those as citations for each statement.

\noindent \textbf{Re-retrieval + NLI rerank}. For each generated statement, we relaunch the retriever to retrieve top-k documents relevant to the statement from $D$, and use a light-weighted medicine NLI model to assign those retrieved documents as citations.

\noindent \textbf{Re-retrieval + LLM rerank}. Similar as above config, except that we instruct LLMs to assign retrieved documents as citations.  

\begin{table}[!ht]
\centering
\newcommand{\smallcolspc}{\hspace*{0.3em}} 
\newcommand{\colspc}{\hspace*{0.3em}}
\tiny
\begin{tabular}{@{\smallcolspc}c@{\smallcolspc}c@{\smallcolspc}c@{\smallcolspc}c@{\smallcolspc}c@{\smallcolspc}c@{\smallcolspc}}
\toprule
\multirow{2}{2cm}{\textbf{Attribution Strategy}} & \multirow{2}{1cm}{\textbf{Accuracy (EM)}} & \multicolumn{2}{c}{\textbf{Text Quality}} & \multicolumn{2}{c}{\textbf{Citation Quality}} \\
\cmidrule(lr){3-4} \cmidrule(lr){5-6}
                        &                        & \textbf{MAUVE} & \textbf{ROUGE-L} & \textbf{Recall} & \textbf{Precision} \\ 
\midrule
\raggedright Pre-Gen. shortlist + LLM rerank & 83.33         & 59.22 &    16.78      & 54.66               & 41.40                     \\ 
\midrule
\raggedright Retriever-only re-retrieval                & 83.33        & 59.22     & 16.78               & 65.69          & 47.69                     \\ 
\midrule
\raggedright Re-retrieval + NLI rerank & 83.33        & 59.22 &    16.78                 & 65.38                     & 55.12                     \\ 
\midrule
\raggedright Re-retrieval + LLM rerank & 83.33        & 59.22 &    16.78              & \textbf{65.78}                     & \textbf{60.95}                    \\ 
\bottomrule
\end{tabular}
\caption{Comparison of citation seeking methods. We use the Hybrid configuration described in Section 4 because it has overall better citation quality.}
\label{tbl:attributor-comparison}
\end{table}

Table~\ref{tbl:attributor-comparison} shows the comparison results of different citation seeking strategies. We find that \emph{re-retrieval + LLM reranking} has the best overall performances, which confirms the benefits of (1) re-retrieval leads to improved citation precision and recall, and (2) citation reranking as an effective approach for seeking high quality citations. It is worth noting that while \emph{re-retrieval + NLI reranking} achieves similar citation recall but with 5.8\% lower recall precision, the NLI model is overall much more lightweight than an LLM. Therefore, if cost is a major constraint, a medicine-specialized NLI classifier can also be considered for citation seeking.

\section{MedCite: A Citation Generation System for LLM-Powered Medical QA}
\label{sec:method}

In the previous section we investigate several important design choices for citation generation of medical tasks. We now aggregate these improvements and evaluate their combined impact and provide it as an open-source framework \name (\fref{fig:overview}). Our final citation generation approach integrates three core components: non-parametric citation (\sref{subsec:parametric-citation}), RAG (\sref{subsec:rag-analysis}), and the retrieval + LLM reranking citation seeking method (\sref{subsec:citation-seeking}). Additionally, we investigate another two important factors that have been under-emphasized in previous work: (1) what if we combine parametric and non-parametric citations through multi-pass approaches;
and (2) the impact of retriever choices to the citation seeking. 

\paragraph{Multi-pass citation generation.} Intuitively, it seems possible to leverage both LLM's internal parametric knowledge to provide initial answer and citations while employing post-generation non-parametric method to validate and refine the citations, utilizing the externally retrieved content. To verify our hypothesis, we consider a new multi-pass method: Similar to the approach in \sref{subsec:rag-analysis}, we employ RAG to generate answers. Different from that approach, we instruct LLM to assign citations to statements based on the retrieved documents while answering the question. Then we retrieve top-k relevant documents to each statement. We deduplicate any redundant citations from these two stages and combine the remaining ones to form the final citations.

Table~\ref{tbl:hybrid-comparison} presents the comparison results of the double-pass method against non-parametric citation. The results indicate that the double-pass approach consistently outperforms the non-parametric method in citation precision and recall while maintaining comparable and slightly better answer correctness. By combining the strengths of both generative and retrieval systems, the double-pass method mitigates the limitations inherent in each individual approach. 

\begin{table}[!ht]
\centering
\newcommand{\smallcolspc}{\hspace*{0.32em}} 
\newcommand{\colspc}{\hspace*{0.32em}}
\scriptsize
\begin{tabular}{@{\smallcolspc}c@{\smallcolspc}c@{\smallcolspc}c@{\smallcolspc}c@{\smallcolspc}c@{\smallcolspc}c@{\smallcolspc}}
\toprule
\multirow{2}{1.5cm}{\textbf{Configuration}} & \multirow{2}{1cm}{\textbf{Accuracy (EM)}} & \multicolumn{2}{c}{\textbf{Text Quality}} & \multicolumn{2}{c}{\textbf{Citation Quality}} \\
\cmidrule(lr){3-4} \cmidrule(lr){5-6}
                &                        & \textbf{MAUVE} & \textbf{ROUGE-L} & \textbf{Rec.} & \textbf{Prec.} \\ 
\midrule
\parbox{2cm}{\centering \textbf{Non-parametric} \\ RAG + Citation Seeker} & 82.85        & 52.22 & 14.79           & 49.01          & 42.77           \\ 
\midrule
\parbox{2cm}{\centering \textbf{Hybrid} \\ Double-pass}   & \textbf{83.33}                     & \textbf{59.22}     & \textbf{16.78}      & \textbf{65.69}          & \textbf{47.69}           \\ 
\bottomrule
\end{tabular}
\caption{Comparison of non-parametric and \name's double-pass method for citation generation.}
\label{tbl:hybrid-comparison}
\end{table}

\paragraph{Hierarchical two-stage ranking based citation retrieval.}
Another factor is the choice of retriever for the citation seeking. The recently proposed MedRAG~\cite{xiong2024benchmarking} uses a Reciprocal Rank Fusion (RRF) based hybrid method to combine results from BM25~\cite{bm25} and MedCPT~\cite{medcpt} to find supporting documents in the pre-generation phase. However, while it is possible to find a broad range of relevant documents to enhance the context of LLM-generated answers, citation retrieval must be more fact-focused to ensure precise and accurate referencing. In the ablation studies, we show that a hierarchical two-stage ranker that first retrieve documents based on key word matching through BM25~\cite{bm25} and then semantic retriever based on MedCPT~\cite{medcpt} brings further improvements in performance in citation quality, validating the importance of the choice of retriever for citation.

\section{Evaluation}
\label{sec:res}

\subsection{Main Results}
\label{sec:main-results}

\begin{table*}[!ht]
\resizebox{\textwidth}{!}{%
\begin{tabular}{c|>{\columncolor{white}}c|>{\columncolor{lime!20}}c>{\columncolor{cyan!20}}c|>{\columncolor{lime!20}}c>{\columncolor{cyan!20}}c|>{\columncolor{lime!20}}c>{\columncolor{cyan!20}}c|>{\columncolor{lime!20}}c>{\columncolor{cyan!20}}c|>{\columncolor{lime!20}}c>{\columncolor{cyan!20}}c|>{\columncolor{lime!20}}c>{\columncolor{cyan!20}}c}
\toprule
\rowcolor{white}
\multirow{2}{*}{\textbf{Model}} & \multirow{2}{*}{\textbf{Method}} & \multicolumn{2}{c|}{\multirow{2}{*}{\textbf{Acc. (EM)}}} & \multicolumn{4}{c|}{\textbf{Text Gen. Quality}} & \multicolumn{6}{c}{\textbf{Citation Quality}} \\ \cmidrule(lr){5-14}
\rowcolor{white}
                                &                   &  &                & \multicolumn{2}{c|}{\textbf{MAUVE}} & \multicolumn{2}{c|}{\textbf{ROUGE-L}} & \multicolumn{2}{c|}{\textbf{Recall}} & \multicolumn{2}{c}{\textbf{Precision}} &\multicolumn{2}{c}{\textbf{F1-Score}}\\ \cmidrule(lr){3-14}
\rowcolor{white}
                                &                                  & \textbf{BioASQ} & \textbf{PubMedQA} & \textbf{BioASQ} & \textbf{PubMedQA} & \textbf{BioASQ} & \textbf{PubMedQA} & \textbf{BioASQ} & \textbf{PubMedQA} &
                                \textbf{BioASQ} & \textbf{PubMedQA} &\textbf{BioASQ} & \textbf{PubMedQA} \\ \midrule

\multirow{4}{*}{Llama-3-8B-I.}     
                                & \cellcolor{orange!10}MedRAG                           & \cellcolor{orange!10}82.85 & \cellcolor{orange!10}\textbf{70.80} & \cellcolor{orange!10}53.74 & \cellcolor{orange!10}42.39 & \cellcolor{orange!10}14.78 & \cellcolor{orange!10}14.22 & \cellcolor{orange!10}/ & \cellcolor{orange!10}/ &
                                \cellcolor{orange!10}/ & \cellcolor{orange!10}/ &
                                \cellcolor{orange!10}/ & \cellcolor{orange!10}/ \\ \cmidrule(lr){2-14} 
                                & \cellcolor{orange!10}PRG                              & \cellcolor{orange!10}\textbf{84.95} & \cellcolor{orange!10}69.40 & \cellcolor{orange!10}\textbf{72.53} & \cellcolor{orange!10}\textbf{47.79} & \cellcolor{orange!10}\textbf{17.97} & \cellcolor{orange!10}\textbf{20.99} & \cellcolor{orange!10}35.44 & \cellcolor{orange!10}30.08 & \cellcolor{orange!10}38.71 & \cellcolor{orange!10}35.00 & \cellcolor{orange!10}32.50 & \cellcolor{orange!10}36.73\\ \cmidrule(lr){2-14} 
                                & \cellcolor{orange!10}PGC*                              & \cellcolor{orange!10}72.10 & \cellcolor{orange!10}55.80 & \cellcolor{orange!10}61.90 & \cellcolor{orange!10}44.53 & \cellcolor{orange!10}18.06 & \cellcolor{orange!10}19.11 & \cellcolor{orange!10}64.75 & \cellcolor{orange!10}62.18 & \cellcolor{orange!10}69.32 &
                                \cellcolor{orange!10}\textbf{71.75} & \cellcolor{orange!10}66.96&
                                \cellcolor{orange!10}66.62 \\ \cmidrule(lr){2-14} 
                                & \cellcolor{orange!10}MedCite                          & \cellcolor{orange!10}\textbf{84.95} & \cellcolor{orange!10}69.40 & \cellcolor{orange!10}\textbf{72.53} & \cellcolor{orange!10}\textbf{47.79} & \cellcolor{orange!10}\textbf{17.97} & \cellcolor{orange!10}\textbf{20.99} & \cellcolor{orange!10}\textbf{74.86} & \cellcolor{orange!10}\textbf{69.50} & \cellcolor{orange!10}\textbf{69.47} & \cellcolor{orange!10}67.73 & \cellcolor{orange!10}\textbf{71.74} & \cellcolor{orange!10}\textbf{68.60} \\ \midrule

\multirow{4}{*}{UltraMedical}    
                                & \cellcolor{cyan!10}MedRAG                           & \cellcolor{cyan!10}\textbf{74.92} & \cellcolor{cyan!10}\textbf{65.00} & \cellcolor{cyan!10}57.24 & \cellcolor{cyan!10}\textbf{58.82} & \cellcolor{cyan!10}\textbf{17.33} & \cellcolor{cyan!10}\textbf{20.54} & \cellcolor{cyan!10}/ & \cellcolor{cyan!10}/ & \cellcolor{cyan!10}/ & \cellcolor{cyan!10}/
                                & \cellcolor{cyan!10}/ & \cellcolor{cyan!10}/ \\ \cmidrule(lr){2-14} 
                                & \cellcolor{cyan!10}PRG                              & \cellcolor{cyan!10}63.43 & \cellcolor{cyan!10}53.60 & \cellcolor{cyan!10}\textbf{63.87} & \cellcolor{cyan!10}48.02 & \cellcolor{cyan!10}13.27 & \cellcolor{cyan!10}14.89 & \cellcolor{cyan!10}27.54 & \cellcolor{cyan!10}28.51 & \cellcolor{cyan!10}30.80 & \cellcolor{cyan!10}31.17 & \cellcolor{cyan!10}28.01 & 
                                \cellcolor{cyan!10}30.94 \\ \cmidrule(lr){2-14} 
                                & \cellcolor{cyan!10}PGC                              & \cellcolor{cyan!10}68.12 & \cellcolor{cyan!10}44.80 & \cellcolor{cyan!10}50.71 & \cellcolor{cyan!10}41.04 & \cellcolor{cyan!10}12.69 & \cellcolor{cyan!10}13.33 & \cellcolor{cyan!10}49.91 & 
                                \cellcolor{cyan!10}54.28 & 
                                \cellcolor{cyan!10}\textbf{62.18} & 
                                \cellcolor{cyan!10}\textbf{72.82} & 
                                \cellcolor{cyan!10}55.37 &
                                \cellcolor{cyan!10}\textbf{62.21} \\ \cmidrule(lr){2-14} 
                                & \cellcolor{cyan!10}MedCite                          & \cellcolor{cyan!10}63.43 & \cellcolor{cyan!10}53.60 & \cellcolor{cyan!10}\textbf{63.87} & \cellcolor{cyan!10}48.02 & \cellcolor{cyan!10}13.27 & \cellcolor{cyan!10}14.89 & \cellcolor{cyan!10}\textbf{74.93} & \cellcolor{cyan!10}\textbf{60.12} & \cellcolor{cyan!10}45.42 & \cellcolor{cyan!10}64.19 & \cellcolor{cyan!10}\textbf{66.71} & \cellcolor{cyan!10}53.14 \\ \midrule

\multirow{4}{*}{GPT-4o}          
                                & \cellcolor{lime!20}MedRAG                           & \cellcolor{lime!20}92.39 & \cellcolor{lime!20}73.80 & \cellcolor{lime!20}51.29 & \cellcolor{lime!20}38.00 & \cellcolor{lime!20}15.77 & \cellcolor{lime!20}24.11 & \cellcolor{lime!20}/ & 
                                \cellcolor{lime!20}/ & 
                                \cellcolor{lime!20}/ & 
                                \cellcolor{lime!20}/ & 
                                \cellcolor{lime!20}/ &
                                \cellcolor{lime!20}/ \\ \cmidrule(lr){2-14} 
                                & \cellcolor{lime!20}PRG                              & \cellcolor{lime!20}\textbf{92.56} & \cellcolor{lime!20}\textbf{75.60} & \cellcolor{lime!20}60.74 & \cellcolor{lime!20}52.32 & \cellcolor{lime!20}19.97 & \cellcolor{lime!20}\textbf{27.18} & \cellcolor{lime!20}53.86 & \cellcolor{lime!20}51.33 & \cellcolor{lime!20}57.27 & \cellcolor{lime!20}55.27 & \cellcolor{lime!20}52.45 & 
                                \cellcolor{lime!20}56.26\\ \cmidrule(lr){2-14} 
                                & \cellcolor{lime!20}PGC                              & \cellcolor{lime!20}87.70 & \cellcolor{lime!20}50.60 & \cellcolor{lime!20}\textbf{67.01} & \cellcolor{lime!20}\textbf{61.72} & \cellcolor{lime!20}\textbf{20.80} & \cellcolor{lime!20}21.37 & \cellcolor{lime!20}79.59 & 
                                \cellcolor{lime!20}75.94 & 
                                \cellcolor{lime!20}81.01 &
                                \cellcolor{lime!20}82.40 & 
                                \cellcolor{lime!20}80.29 &
                                \cellcolor{lime!20}79.04 \\ \cmidrule(lr){2-14}
                                & \cellcolor{lime!20}MedCite                          & \cellcolor{lime!20}\textbf{92.56} & \cellcolor{lime!20}\textbf{75.60} & \cellcolor{lime!20}60.74 & \cellcolor{lime!20}52.32 & \cellcolor{lime!20}19.97 & \cellcolor{lime!20}\textbf{27.18} & \cellcolor{lime!20}\textbf{84.86} & \cellcolor{lime!20}\textbf{84.54} & \cellcolor{lime!20}\textbf{83.85} & \cellcolor{lime!20}\textbf{89.43} & \cellcolor{lime!20}\textbf{84.36} & \cellcolor{lime!20}\textbf{86.48}\\ \bottomrule
\end{tabular}%
}

\caption{Comparison results of MedCite and alternative methods on BioASQ and PubMedQA datasets. * The generation phase for PGC utilizes CoT, which is non-RAG. Consequently, the Accuracy (EM) score for PGC is the same as that of the CoT (non-RAG) method.}
\label{tbl:main-results}
\end{table*}

We compare \name with three baseline methods: the medical domain RAG method and two general-domain citation methods from recent work, including the post-retrieval generation and post-generation citation method across different backbone LLMs: 

(1) \textbf{MedRAG}: The method described in \cite{xiong2024benchmarking}.

(2) \textbf{Post-retrieval generation (PRG)}: Following the method in~\cite{alce}, we prompt LLMs with a query, a list of retrieved documents and instruct the LLMs to include citations in their generated answer.

(3) \textbf{Post-generation citation (PGC)}: Following RARR~\cite{gao2023rarrresearchingrevisinglanguage}, We perform chain-of-thought (CoT) prompting~\cite{cot} to let LLM generate an answer, followed by the re-retrieval + LLM reranking to assign citations to each statement.

We evaluate three models: Llama-3-8B-I. (Llama-3-8B-Instruct)~\cite{llama3}, UltraMedical~\cite{ultra-medical}, and commercial LLM GPT-4o (gpt-4o-0806)~\cite{gpt4o}. 

We present the main results in Table~\ref{tbl:main-results}. The main takeaways from the experiments are as follow.

\paragraph{Generated responses remain correct with enabled citations.} State-of-the-art medical QA systems such as MedRAG do not have citations in their generated answers. We show that it is possible to enable citations in medical systems while maintaining the correctness of generated answers. In particular, both \name and PRG are able to achieve comparable accuracy, MAUVE, and ROUGE scores to MedRAG on Llama-3-8B-I. and GPT-4o while providing citations to support generated answers. On the other hand, UltraMedical obtains the highest accuracy with MedRAG despite with an absolute accuracy (74.92\%) much lower than Llama-3-8B-I. (82.85\%) and GPT-4o (92.39\%). 
By examining the generated output from UltraMedical, we find that adding additional instructions seems to confuse the model, leading to incorrect responses. This can be because UltraMedical was trained with a context length of 2048, making it harder for the model to focus on the most relevant parts of the prompt as additional instructions are provided.  

\paragraph{\name outperforms PRG and PGC in citation quality.} While both PRG and PGC enable citation for medicine, \name outperforms the two methods by a large margin (e.g., 71.74\% vs. 66.96\% and 32.50\% on BioASQ). \name outperforms PRG because \name's second pass of citation seeking leverages post-generation non-parametric retrieval to refine the citations, which allows LLMs to mitigate citation hallucinations. \name obtains better performance than PGC, because it exploits pre-generation retrieval and LLM's internal parametric knowledge to obtain an initial set of citations, which turns out to be useful for obtaining high-quality final citations. These results have demonstrated \name's effectiveness in combining the strengths of both generative and retrieval systems for citation generation.

\paragraph{\name consistently brings citation quality improvements over different LLMs.} We see a universal trend that \name improves citation recall and F$_1$ score across LLMs. 
Using GPT-4o as the backbone LLM leads to the highest-performing citation quality (e.g., GPT-4o 86.48 vs. Llama-3-I. 68.60 in F$_1$ on PubMedQA), mainly driven by its advanced reasoning and instruction following capabilities. In contrast, citation quality is the lowest when the system is evaluated on UltraMedical (e.g., 66.71 on BioASQ). These results underscore that incorporating \name bolsters LLM's capacity to generate verifiable texts.

\paragraph{Qualitative case study.}A qualitative case study is presented in Appendix~\ref{appendix:case-study}, providing concrete examples that complement our quantitative findings.

\subsection{Ablation Studies}

\paragraph{Citation retrieval analysis.}
We evaluate how different citation retrievers affect the quality of \name.  In particular, we compare semantic-only~\cite{medcpt}, lexical-only~\cite{bm25}, and retrieval-fusion via RRF-2~\cite{xiong2024benchmarking}, and hierarchical two-stage retriever. Different from prior findings that the RRF-2 based hybrid retriever leads to the best performance results, we find that lexical-only (e.g., BM25) retriever leads to the higher citation quality. Unlike the retriever used in RAG, which aims to provide supporting documents for LLM generation, citation retrieval requires examination of precise medical terminology and quoting verbatim from the source. For example, in our experiment, given the LLM claim "peptides are short chains of amino acids, and chlorotoxin is a specific type of peptide," the semantic retriever retrieves a document discussing the features of calitoxin. Although both calitoxin and chlorotoxin are toxins, the document does not help support the claim. Therefore, it cannot serve as a valid citation for this statement.
Because of this, a lexical retriever based on exact match provides more precise citations. In contrast, semantic-only and retrieval-fusion based retrievers negatively affect the citation quality. Finally, the hierarchical two-stage retriever fist performs lexical retrieval to obtain a long list of citation candidates followed by a semantic retriever to rank the long list by the similarity score between the query and the citation candidates. As a result, it offers the best-performing results among our tested configurations by achieving a good trade-off between citing comprehensively and precisely. 

\begin{table}[!ht]
\centering
\resizebox{\columnwidth}{!}{%
\begin{tabular}{@{}lcccc@{}} 
\toprule
\multirow{2}{*}{\textbf{Retriever Type}} & \multirow{2}{*}{\textbf{Method}} & \textbf{Accuracy} & \multicolumn{2}{c}{\textbf{Citation Quality}} \\ 
\cmidrule(lr){4-5}
                        &                  & \textbf{(EM)}    & \textbf{Rec.} & \textbf{Prec.} \\ 
\midrule
Lexical-only           & BM25            & 94.34                    & 77.53             & 79.89            \\ 
\midrule
Semantic-only          & MedCPT          & 94.34                    & 65.93             & 66.78            \\ 
\midrule
Combination            & RRF-2           & 94.34                    & 75.74             & 76.46            \\ 
\midrule
Hierarchical            & BM25 then MedCPT           & 94.34                    & \textbf{77.84}             & \textbf{80.02}            \\ 
\bottomrule
\end{tabular}%
}
\caption{Effectiveness of different retrievers on MedCite quality with Llama-3-8B-I. {using Oracle relevant documents as the supporting documents in the pre-generation retrieval stage and re-retrieve top-3 documents per statement with LLM reranking.}}
\label{tbl:retriever-effectiveness}
\end{table}

\subsection{Attribution judge analysis and human annotations.} 
While prior studies often assume that Natural Language Inference (NLI) models correlate well with human judgements in making attribution evaluation~\cite{alce,attributed-qa}, those studies focus on general domain questions. To our knowledge, no study has evaluated the effectiveness of different models in attribution judgment for medical tasks. We evaluate models for medical attribution and compare them to medical expert judgments. Annotation guidelines are provided in Appendix~\ref{appendix:annotataion-guide}.

\label{para:human_eval}
Surprisingly, Table~\ref{table:correlation} indicates that existing medicine-specialized NLI models exhibit poor correlation with professional medical doctor judgments (e.g., $<$22.3\% score in precision judge). Also interestingly, GPT-4o/GPT-3.5 are not the top performing models in this context. Instead, public models such as Llama-3.1 and Mistral achieve the best correlation with expert judgments, demonstrating a higher level of agreement with medical professionals. We hypothesize that this could be because public LLMs might have been trained on datasets that include more medical literature, although it is hard to verify because the details of the datasets used for training these models are not publicly disclosed. At the same time, we recognize that reasoning capabilities play a central role in the attribution judgment task, as described in Appendix \ref{subsec:prompts}, the prompt requires models to evaluate the connection between a premise and a hypothesis based on self-contained excerpts. However, domain knowledge remains essential for interpreting specialized claims\cite{wadden-etal-2020-fact}. For instance, verifying claims such as “Cardiac injury is common in critical cases of COVID-19” requires medical expertise to connect elevated troponin levels with cardiac injury. Thus, better-performing LLMs likely benefit from extensive pretraining on medical datasets, which enhances both reasoning and domain-specific understanding. Nevertheless, given the high correlation between recent top-performing LLMs and expert judgments, we consider using LLMs as attribution judgements to be more promising for medicine, and we see this as an opportunity for future work. 

Expert annotations can vary in knowledge-intensive domains like medicine. In the SciFact dataset, inter-annotator agreement (Cohen’s kappa) is about 0.75~\cite{phan2021scifivetexttotexttransformermodel}. In our study, we observed similar agreement: 0.83 for statement-level recall and 0.66 for citation-level precision, reflecting consistency comparable to prior work despite the task complexity.

\begin{table}[]
\resizebox{\columnwidth}{!}{%
\begin{tabular}{lllcc}
\hline
\multirow{2}{*}{\textbf{Model}} & \multirow{2}{*}{\textbf{Source}} & \multirow{2}{*}{\textbf{Domain}} & \multicolumn{2}{l}{\textbf{Cohen's Kappa Score}} \\ \cline{4-5} 
 &  &  & \textbf{\begin{tabular}[c]{@{}c@{}}Rec.\\ Judge\end{tabular}} & \textbf{\begin{tabular}[c]{@{}c@{}}Prec.\\ Judge\end{tabular}} \\ \hline
SciFive-MedNLI & Open & Medical & 0.2593 & 0.1945 \\ \hline
JSL-MedPhi2-2.7B & Open & Medical & 0.1845 & 0.2218 \\ \hline
UltraMedical & Open & Medical & 0.4518 & 0.2162 \\ \hline
Llama-3.1-8B-Instruct & Open & General & 0.5862 & \textbf{0.5422} \\ \hline
mistral-7B-Instruct & Open & General & \textbf{0.6211} & 0.4241 \\ \hline
GPT-3.5-Turbo & Close & General & 0.3834 & 0.4075 \\ \hline
GPT-4o & Close & General & 0.4146 & 0.4075 \\ \hline
GPT-4o-mini & Close & General & 0.3834 & 0.3894 \\ \hline
\end{tabular}
}
\caption{Correlation of different models' attribution judge with human annotations.}
\label{table:correlation}
\end{table}

\section{Conclusion}
\label{sec:conclusion}

We introduce \name, the first end-to-end framework fostering research that targets improving the verifiability and trustworthy of medical systems with citations. Our in-depth examination of important design choices for LLM-based medical systems inspires us to propose \name, a novel method for generating high quality citations for medical systems. 

Extensive evaluation across LLMs show that our approach leads to consistent improvements to citation generation over alternative methods.


\section{Ethical Considerations and Limitations}
\label{sec:limitations}

The primary goal of this work is to assess and improve the verifiability in LLM-based medical systems via citations. In addition to gaining trust from physicians and patients, there is also urgency from regulation and audition consideration, where the US Food and Drug Administration (FDA) has called for regulation methods for using LLMs in the medical industry~\cite{fda-regulation}. 
However, incorrect citations can have serious consequences in the medical field, as they can affect patient and physician's treatment decisions. As such, the deployment of LLM-based systems in medical contexts requires careful design while adhering to ethical considerations, e.g., the system should augment human decision-making rather than replace it, and human oversight remains critical to validate generated citations.

One of the limitation of this work is that we did not perform extensive hyperparameter tuning (e.g., the number of retrieved passages), instead following prior empirical findings and practical constraints. While our choices are justifiable, further tuning could potentially improve performance.

While conducting this research, we have also identified several critical yet unexplored challenges in generating citations for medicine. For instance, manual human verification by professional medical doctors remains a costly and time-intensive process, making it difficult to scale. Additionally, whether a document supports a statement can be subject to interpretation, even among medical experts, who may disagree on the extent to which a document partially supports a statement. Therefore, it is crucial to assess whether a high level of consensus among doctors can be achieved. Another challenge is the limited availability of medical datasets that include both ground truth answers and supporting documents, aside from BioASQ and PubMedQA. The absence of certain information, such as ground truth references, in medical datasets complicates the overall verifiability evaluation in medicine. Future work should focus on developing high-quality citation datasets for medicine, which would significantly enhance the trustworthiness and effectiveness of medical QA systems, ultimately benefiting healthcare professionals and patients. Similarly, although we did not observe many such cases in our study, it is worth exploring the reasons behind instances where citations disprove claims. This is particularly relevant when considering corpus updates, as new research may refute prior studies, raising important questions about how to handle such scenarios effectively.

While MedCite is specifically tailored to the medical domain, its generalization to other fields presents notable challenges. Key components, such as multi-pass citation generation and curated database reliance, may not translate directly to general-domain applications without significant modifications. For example, the availability of well-curated corpora like PubMed is unique to the medical field. General domains often lack centralized resources, requiring extensive dataset preparation or integration of diverse sources. Similarly, retriever selection, such as the use of MedCPT in this study, may need to be adapted to align with the characteristics and retrieval objectives of different fields. The effectiveness of retrieval configurations and strategies could vary significantly depending on corpus diversity and domain-specific needs. Moreover, citation evaluation strategies may need to accommodate varying requirements across domains. In medicine, most claims necessitate citations due to high stakes and reliance on specialized knowledge, whereas general domains may involve claims rooted in common sense or widely accepted facts. Evaluating citations in such contexts might require adjustments to account for optional citations or more loosely defined relevance criteria. Automatic evaluation approaches, while valuable, would also need adaptation to handle the simpler or binary relationships typical of claims and citations in general fields. These limitations suggest that while MedCite’s core framework offers a strong foundation, further work is needed to ensure its components are broadly applicable to non-medical domains.

\section*{Acknowledgments}

We sincerely appreciate the insightful feedback from the anonymous reviewers. This research was supported by the National Science Foundation (NSF) under Grant No. 2441601. The work utilized the DeltaAI system at the National Center for Supercomputing Applications (NCSA) through allocation CIS240055 from the Advanced Cyberinfrastructure Coordination Ecosystem: Services \& Support (ACCESS) program, which is supported by National Science Foundation grants \#2138259, \#2138286, \#2138307, \#2137603, and \#2138296. The Delta advanced computing resource is a collaborative effort between the University of Illinois Urbana-Champaign and NCSA, supported by the NSF (award OAC 2005572) and the State of Illinois. This work also utilized the Illinois Campus Cluster and NCSA NFI Hydro cluster, both supported by the University of Illinois Urbana-Champaign and the University of Illinois System. This research was supported in part by the Division of Intramural Research (DIR) of the National Library of Medicine (NLM), National Institutes of Health.

\bibliography{reference}

\appendix

\section{Related Work}
\label{sec:related}

\subsection{Biomedical Question Answering}
Biomedical question answering (QA) is a specialized field within natural language processing. It focuses on answering questions related to biomedical and clinical domains. Early approaches rely heavily on rule-based systems~\cite{medqa,askhermes}. These methods utilize structured databases and ontologies to retrieve answers to clinical questions. While pioneering, these systems were limited by their reliance on predefined rules and lack of scalability. Subsequently, ML/DL based solutions have brought significant improvements to biomedical QA. Models such as BioBERT~\cite{biobert} and ClinicalBERT~\cite{clinical-bert} adapt pre-trained BERT~\cite{bert} to biomedical texts, resulting in improved performance on various biomedical QA tasks~\cite{llm-bioqa}. Recently, generative models represent a newer paradigm in biomedical QA. Models such as GPT-3.5/4~\cite{gpt-3,gpt4} and Med-Gemini~\cite{med-gemini} generate answers directly from input text without relying on predefined answer options, which enable more flexible and contextually appropriate responses. However, generative models also pose challenges, such as the risk of generating incorrect or hallucinated answers. To tackle the issue, recent work employs retrieval-augmented generation (RAG) to retrieve relevant documents and generate answers based on the retrieved information~\cite{lozano2023clinfo,xiong2024benchmarking,llm-bioqa,enhance-healthcare,zakka2024almanac,i-medrag}. Different from these efforts, we focus on improving the verifiability of medical systems.

\subsection{Citation Methods for LLM Generation}

The integration of citation mechanisms in LLM based generation is a burgeoning area of research. Recent advancements in LLMs can be prompted to include citations in the text it generates~\cite{gpt-3,lamda,gemini-google,gpt4,gpt4o}. However, the accuracy and relevance of these citations can be a challenge. Similar as hallucination in generated texts, the model (e.g., ChatGPT) can generate plausible-looking citations that are not actually accurate or verifiable~\cite{chatgpt-attribution-hallucination}. 
Multiple methods have been proposed to add citations to LLM-generated content. Direct model-driven attribution methods allow the model to self-attribute, though this often leads to unreliable results~\cite{recitation-augmented,agrawal2023language,according-to}. Post-retrieval generation (PRG) involves retrieving a list of documents relevant to the user query before generating an answer and the relevant documents~\cite{rag-pretraining,retrieval-trillions-tokens,reddy2023smartbook}. 
Post-generation citation (PGC) seeks relevant documents after generating the answer~\cite{retrieve-supporting-evidence}. Both PRG and PGC offer more reliable attribution but increase system complexity~\cite{gao2023rarr}, and as we show in the paper, they may not achieve the optimal citation quality for medicine systems due to the nuanced nature of biomedical queries and the need for precise, verifiable citations. Our hybrid double-pass citation method aims to address these gaps by integrating RAG with post-generation refinement.
Fine-tuning LLMs for citation generation represents another approach, where models are trained using curated or synthetic data to directly produce citations during text generation~\cite{ye2024effectivelargelanguagemodel}.
Finally, there has been overall an absence of automated evaluation for the citation methods over LLM-based QA. Therefore, there has been efforts that aim to improve the evaluation protocols and benchmarks for LLM attributions~\cite{measuring-nlg-attribution,alce,towards-verifiable-generation}. Different from those efforts, which measures citations for general domain subjects, our evaluation is medicine-centric and we also explore the other components, such as medical-specific retrieval and citation seeking  strategies that impact LLM based medicine tasks.

\subsection{Evaluation Frameworks for LLM-Generated Citations in Medical Domain}
\citet{wu2024llmsciterelevantmedical} introduced an evaluation pipeline for assessing the validity of LLM-generated citations in medical domain, focusing on URL-based online sources. Their work highlights significant limitations in LLM citation quality, with even top-performing models like GPT-4 (RAG) failing to fully support all statements in nearly half of their responses.

While both our study and Wu et al. share the goal of improving citation reliability, our work differs in scope and methodology. Wu et al. provides a comprehensive evaluation pipeline, with primary focus on analyzing citation quality for parametric methods by prompting API-based LLMs to provide source URL in their answer, rather than proposing methods to address identified gaps. In contrast, our work not only evaluates but also introduces a modular framework combining hierarchical retrieval and multi-pass citation to improve citation quality for biomedical tasks. In Section 3.1, we explain why parametric methods are unsuitable especially for open-source LLMs due to challenges such as fabricated citations, lack of access to reliable content, and the difficulty of automatic evaluation without API-level access to online sources. By emphasizing domain-specific hierarchical retrieval from curated medical sources like PubMed, we address challenges unique to the biomedical domain, such as ensuring precision for highly specialized terms like drug names or genomic markers.

\section{Additional Experimentation Details}
\label{appendix:extra-exp-details}

\subsection{Datasets}
\label{appendix:dataset}

We use medical question answering datasets that have ground truth answers to evaluate \name.  Specially, we use BioASQ~\cite{bioasq} and PubMedQA~\cite{pubmedqa} in the final evaluation. In both cases, we only use questions and remove all ground truth supporting contexts, which represents a more realistic setting as often no demonstrations are provided in real usage scenarios. Table~\ref{tbl:dataset-overview} summarizes the details about these two datasets.

\begin{table}[!ht]
\centering
\resizebox{\columnwidth}{!}{%
\begin{tabular}{@{}lcccc@{}} 
\toprule
\textbf{Dataset}      & \textbf{Size} & \textbf{Question Type} & \textbf{Example Question} & \textbf{GT Answer} \\ 
\midrule
\textbf{PubMedQA*}    & 500           & Yes/No/Maybe           & \parbox{5cm}{Is anorectal endosonography valuable ... ?}        & yes                  \\ 
\midrule
\textbf{BioASQ-Y/N}   & 618           & Yes/No                 & \parbox{5cm}{Is medical hydrology the same as Spa ... ?}        & yes                  \\ 
\bottomrule
\end{tabular}%
}
\caption{The two datasets used in MedCite experiments.}
\label{tbl:dataset-overview}
\end{table}

\paragraph{PubMedQA.} PubMedQA is a dataset for biomedical question answering (QA) tasks. The questions are either the titles of existing research articles or derived from them. The context provides the abstract of the article. The answer includes a ground truth answer to the question, which is derived from the conclusion of the abstract.

\paragraph{BioASQ-Y/N.} BioASQ-Y/N is also a biomedical QA dataset. For each instance in the dataset, it contains a question, contexts that provide the information to answer the question, and human annotated answers.

\subsection{Hyperparameters}
\label{appendix:hparams}
To ensure reproducibility, we use greedy decoding for all LLMs. For retrieval, we use a hierarchical two-stage ranking process: (1) BM25 implemented with Pyserini~\cite{10.1145/3404835.3463238} using default hyperparameters for indexing, and (2) MedCPT Cross-Encoder\footnote{https://huggingface.co/ncbi/MedCPT-Cross-Encoder} with default settings to rank the retrieved documents for a given query. We retrieve the top-32 documents for answer generation, ensuring they fit within the model's context window, and discard those with lower similarity scores if necessary. We retrieve top-3 documents for a single statement when seeking citation after answer generation. We chose $k=3$ for retrieved documents based on two main considerations: (1) the context length limit of UltraMedical (1024 tokens), which accommodates approximately three PubMed abstracts, making $k=3$ a natural fit across all models for fair comparison; (2) prior analysis by Gao et al.~\cite{gao2023rarr} showed that citation quality plateaus at top-3 passages, further supporting our choice.

\subsection{Correlation between Rouge-L and Accuracy}

Table \ref{table:relation_rouge_acc} illustrates the relationship between ROUGE scores and accuracy under different conditions:

\begin{table}[H]
\centering
\resizebox{\columnwidth}{!}{
\begin{tabular}{|c|c|c|}
\hline
\textbf{Conditions} & \textbf{ROUGE-L Score} & \textbf{Accuracy} \\ 
\hline
medrag + medcpt & 17.04 & 0.8414 \\
medrag + MedCPT + system prompt & 17.98 & 0.8576 \\
medrag + medcpt + new prompt & 17.34 & 0.8269 \\
oracle relevant docs & 22.00 & 0.9401 \\
\hline
\end{tabular}
}
\caption{Analysis of ROUGE-L Scores and Accuracy under Different Conditions}
\label{table:relation_rouge_acc}
\end{table}

The table clearly demonstrates a positive correlation between ROUGE scores and accuracy. Specifically, when the system prompt is introduced to the MEdRAG model, the ROUGE score increases from 17.04 to 17.98, and the accuracy also improves from 0.8414 to 0.8576. This indicates that by optimizing the prompts, we can enhance the model's output quality and accuracy to some extent. Moreover, when a new prompt is introduced, although the ROUGE score slightly decreases, the accuracy drops more notably, suggesting that the new prompt may have affected the model's performance in certain aspects. Most notably, when using the oracle relevant documents, both the ROUGE score and accuracy reach their peak values, further confirming the positive correlation between ROUGE scores and the accuracy of the model's output. These results suggest that ROUGE scores can serve as an effective metric to assess and optimize the output quality of Large Language Models (LLMs).

\section{Prompt Templates}
\label{subsec:prompts}

\begin{figure}[!h]
    \centering
    \includegraphics[width=\columnwidth]{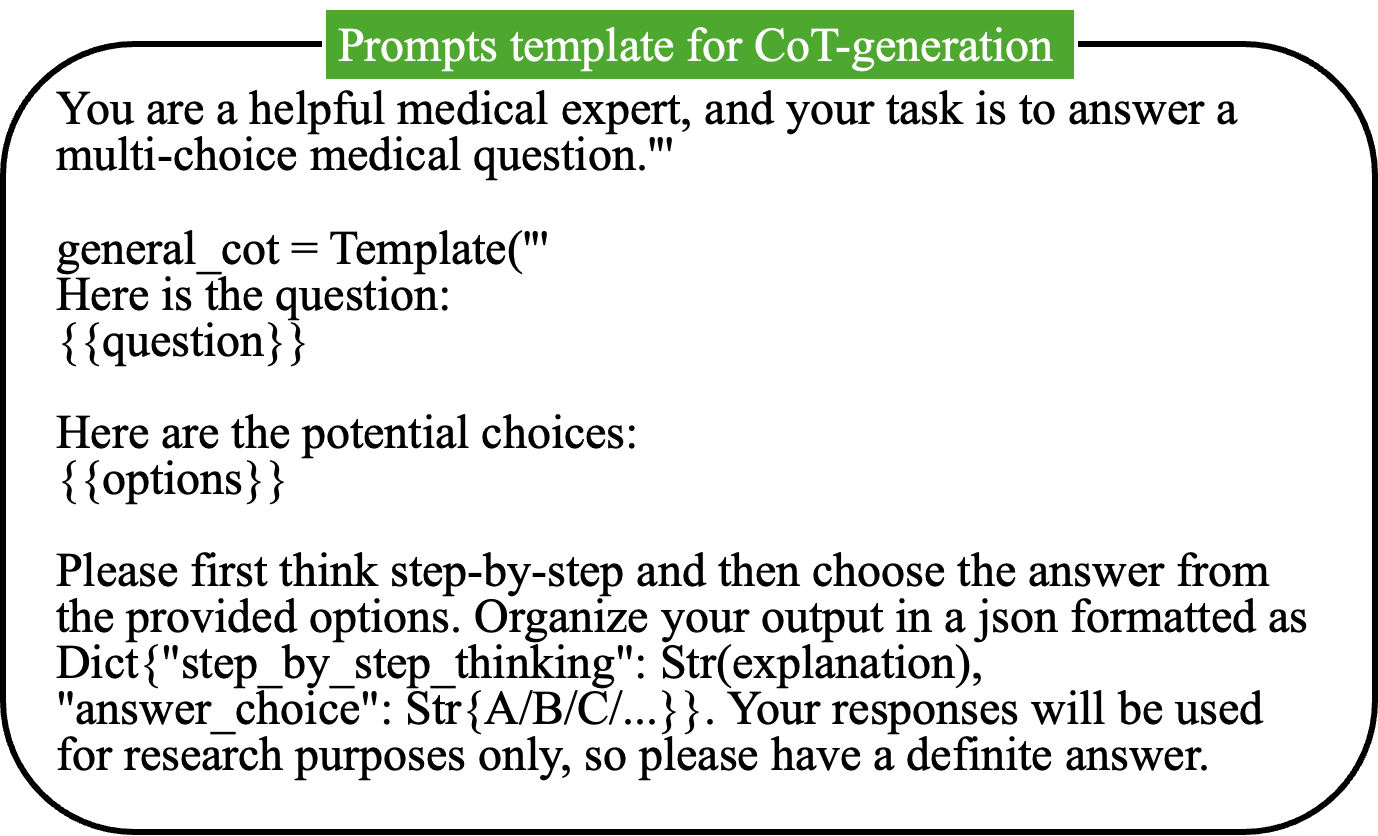}
    \caption{Prompt templates used for CoT generations.}
    \label{fig:motivation}
\end{figure}

\begin{figure}[!h]
    \centering
    \includegraphics[width=\columnwidth]{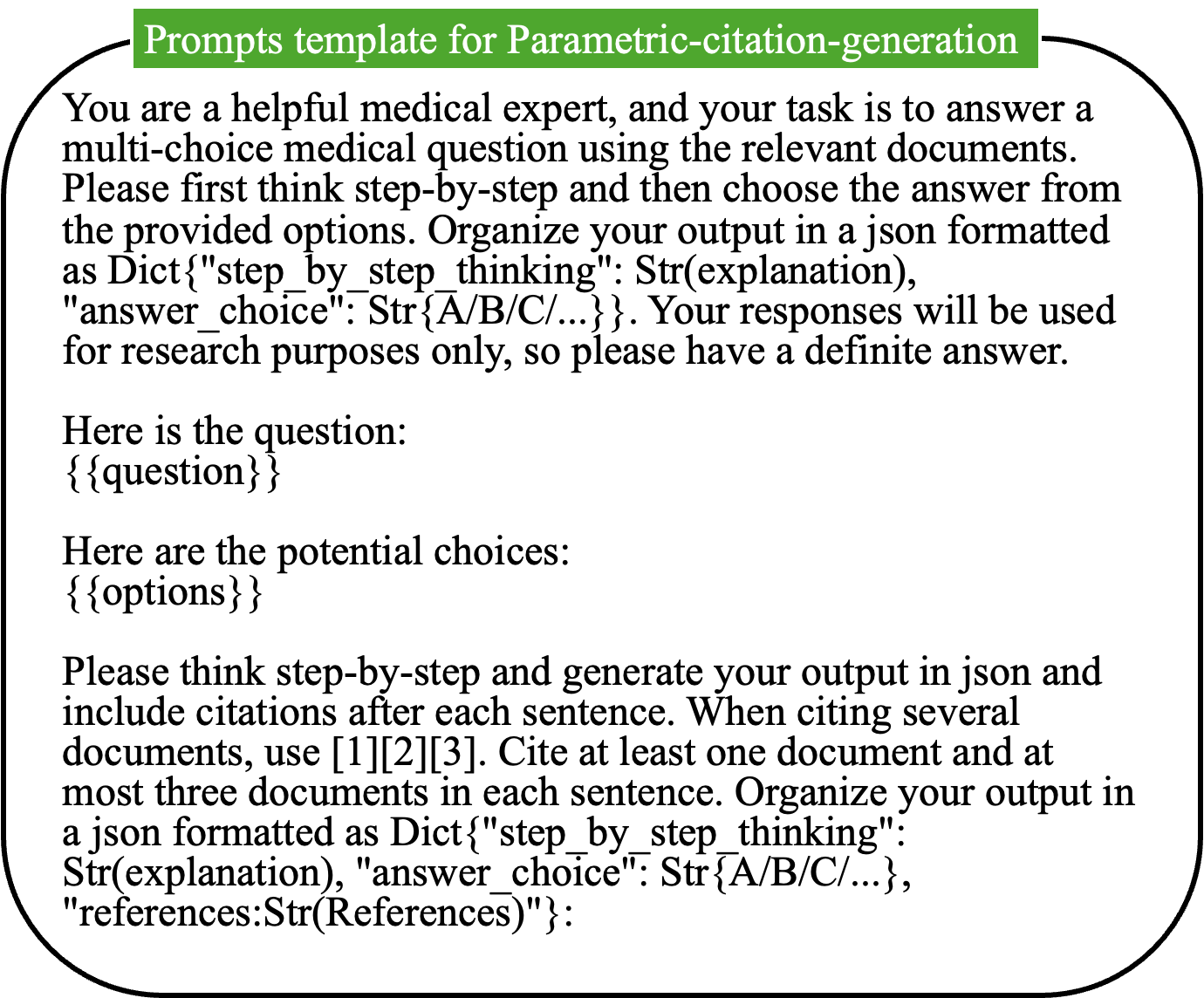}
    \caption{Prompt templates for parametric citations.}
    \label{fig:motivation}
\end{figure}

\begin{figure}[!h]
    \centering
    \includegraphics[width=\columnwidth]{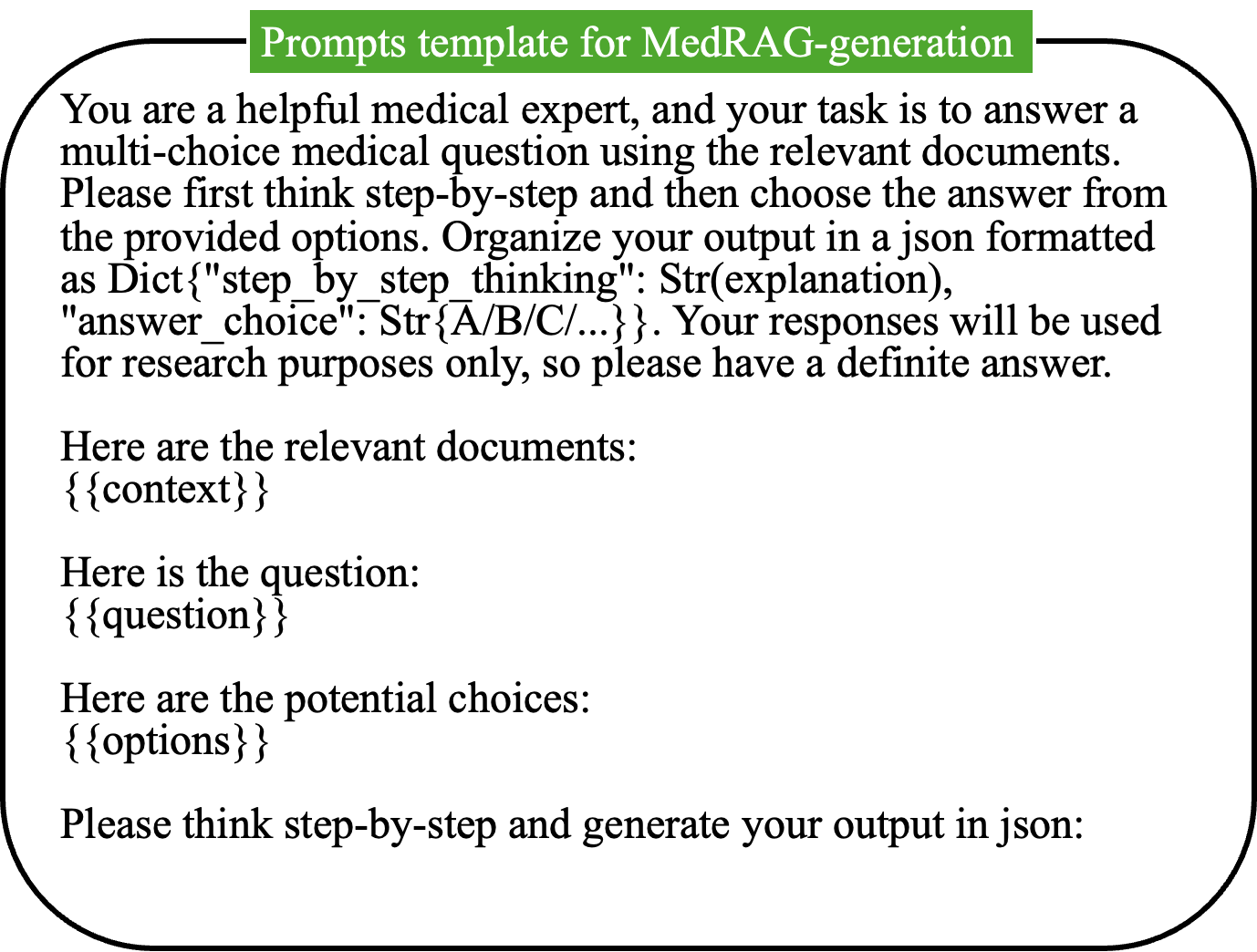}
    \caption{Prompt templates for MedRAG generations.}
    \label{fig:motivation}
\end{figure}

\begin{figure}[!h]
    \centering
    \includegraphics[width=\columnwidth]{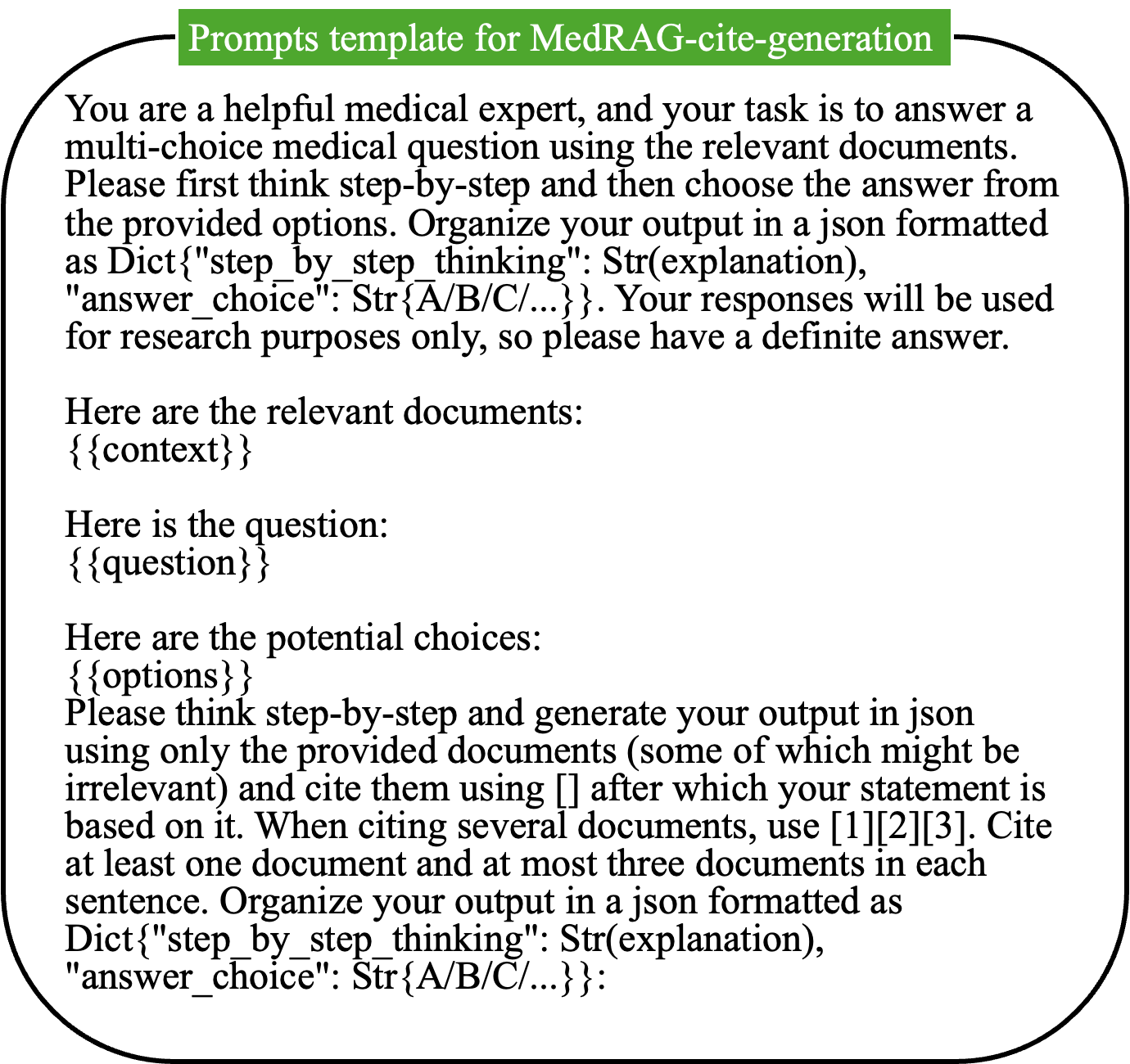}
    \caption{Prompt templates for MedRAG plus citation generations.}
    \label{fig:motivation}
\end{figure}

\begin{figure}[!h]
    \centering
    \includegraphics[width=\columnwidth]{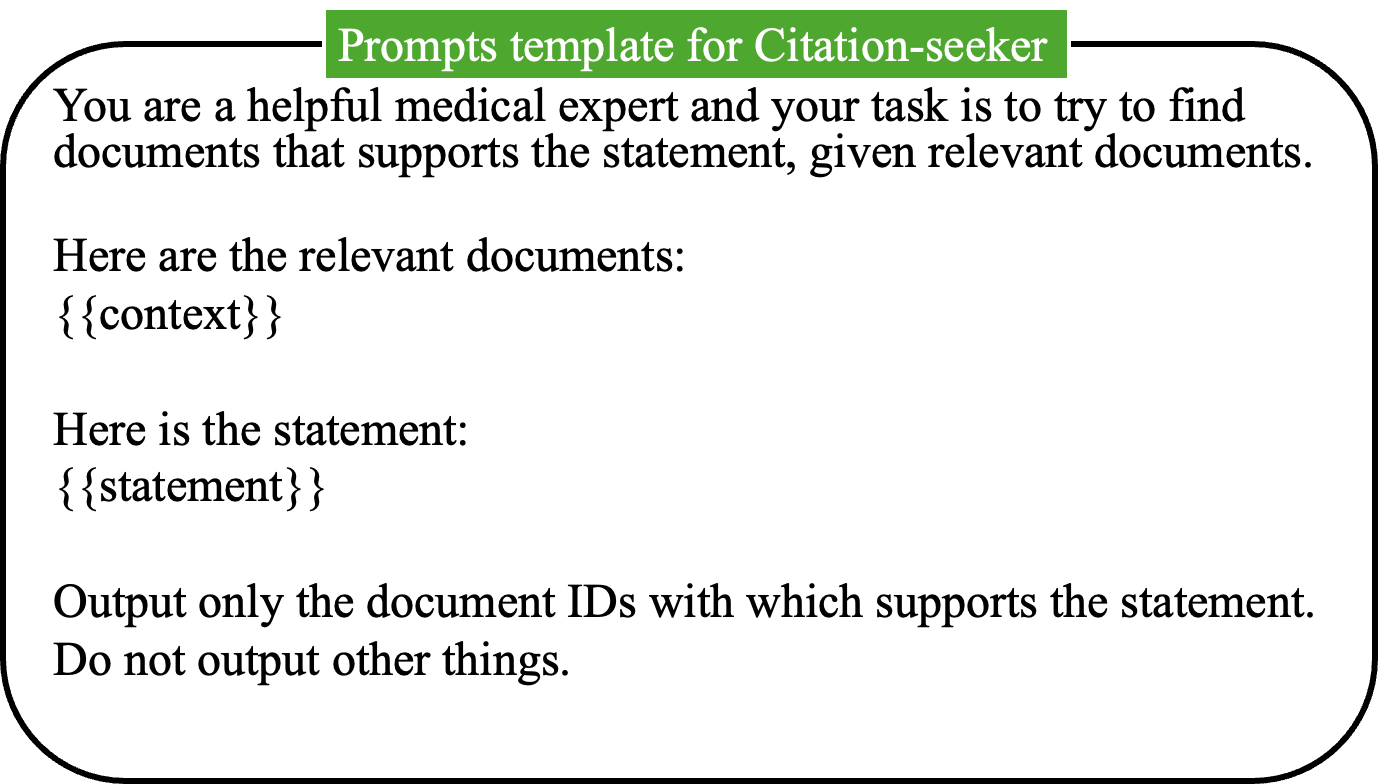}
    \caption{Prompt templates for citation seekers.}
    \label{fig:motivation}
\end{figure}

\begin{figure}[!h]
    \centering
    \includegraphics[width=\columnwidth]{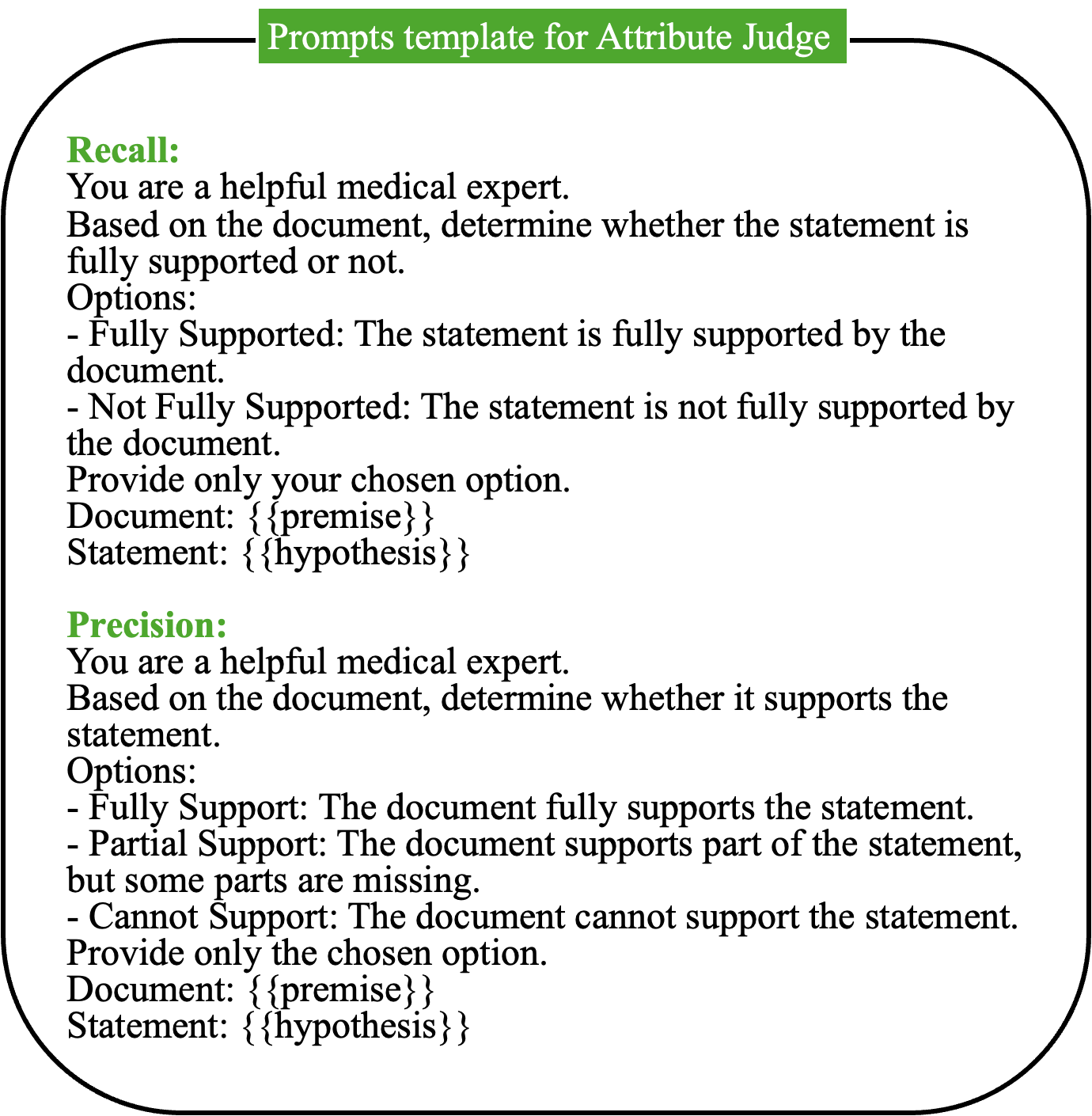}
    \caption{Prompt templates for attribute judge.}
    \label{fig:motivation}
\end{figure}


\section{Examples of Generated Citations}
\label{subsec:ill-formated-citations}

Table~\ref{table:generated-citation-example} shows examples of generated medical references with parametric citation method using Llama-3-8B-I., UltraMedical, and GPT-4o. For Llama-3-8B-I., the URL provided in Reference [1] is incorrect, and References [2] and [3] have different authors despite having the same title. Upon inspection, it was found that the article in question does not exist. UltraMedical includes poorly formatted in-line citations and fabricated references. GPT-4o, on the other hand, provides correct references, but they are challenging to evaluate due to limited API access to the sources.

\begin{table*}[!ht]
\centering
\begin{tabular}{p{16cm}}
\hline\hline
\textbf{Question:} Is obesity related to cognitive decline? \\
\hline
\textbf{Llama-3-8B-I. answer:} \\
Obesity is a complex condition that is associated with various health risks, including cardiovascular disease, type 2 diabetes, and certain types of cancer [1]. Research has also suggested that obesity may be linked to cognitive decline, as it can lead to chronic inflammation, oxidative stress, and changes in brain structure and function [2]. Studies have shown that obese individuals are at a higher risk of developing dementia and Alzheimer's disease, and that weight loss may help to improve cognitive function [3]. Therefore, it is likely that obesity is related to cognitive decline. \\
\textbf{References}: 
[1] World Health Organization. (2018). Obesity and overweight. Retrieved from <https://www.who.int/news-room/fact-sheets/detail/obesity-and-overweight>; 
[2] Yaffe, K., et al. (2013). The association between obesity and cognitive decline in older adults. Journal of the American Medical Association, 310(11), 1155-1162; 
[3] Luchsinger, J. A., et al. (2013). Obesity and cognitive decline in older adults. Journal of Alzheimer's Disease, 35(2), 247-255. \\
\hline
\textbf{UltraMedical answer:} \\
Obesity is a condition characterized by excessive body fat that poses a risk to health. It is associated with various health issues, including cardiovascular diseases, diabetes, and certain types of cancer. To determine if obesity is related to cognitive decline, we need to consider the evidence linking obesity to brain health. 1. Research has shown that obesity can lead to changes in the brain's structure and function. For instance, studies have indicated that obesity is associated with reduced brain volume and alterations in brain activity, which could potentially affect cognitive functions. 2. Additionally, obesity is known to increase the risk of metabolic syndrome, which includes conditions such as hypertension, high blood sugar, and high cholesterol. These conditions have been linked to an increased risk of cognitive decline and dementia. 3. Inflammation is another factor to consider. Obesity is a state of chronic low-grade inflammation, and inflammation is a known risk factor for cognitive decline. Adipose tissue, particularly visceral fat, secretes pro-inflammatory cytokines that can affect brain health. 4. Furthermore, obesity can lead to sleep apnea and other sleep disturbances, which are associated with cognitive deficits. Poor sleep quality has been linked to a higher risk of cognitive decline and Alzheimer's disease. 5. Based on the evidence, it is reasonable to conclude that obesity is related to cognitive decline. The mechanisms include direct effects on brain structure and function, metabolic disturbances, inflammation, and sleep disturbances. Therefore, the answer is A. Yes, obesity is related to cognitive decline. \\
\textbf{References}: 
[1] Hall, K. D., \& Guo, J. (2017). Obesity: portable operating instructions. JAMA, 318(17), 1685-1686; 
[2] Yaffe, K., et al. (2013). The metabolic syndrome, inflammation, and risk of cognitive decline and dementia: the Health, Aging, and Body Composition Study. JAMA, 310(3), 197-208; 
[3] Spiegel, K., et al. (2009). Sleep, cognition, and health: emerging research directions. Dialogues in Clinical Neuroscience, 11(4), 667-679. \\
\hline
\textbf{GPT-4o answer:} \\
Obesity has been linked to cognitive decline through various mechanisms. Excess body fat can lead to inflammation and insulin resistance, both of which are associated with negative effects on brain health [1][2]. Additionally, obesity is a risk factor for cardiovascular diseases, which can impair blood flow to the brain and contribute to cognitive decline [2][3]. Studies have shown that individuals with obesity may have a higher risk of developing conditions like Alzheimer's disease and other forms of dementia [1][3]. Therefore, there is substantial evidence supporting the relationship between obesity and cognitive decline. \\
\textbf{References}: 
[1] Smith, E., Hay, P., Campbell, L., \& Trollor, J. N. (2011). A review of the association between obesity and cognitive function across the lifespan: implications for novel approaches to prevention and treatment. Obesity Reviews, 12(9), 740-755; 
[2] Kivipelto, M., Ngandu, T., Fratiglioni, L., Viitanen, M., Kreholt, I., Winblad, B., \& Nissinen, A. (2005). Obesity and vascular risk factors at midlife and the risk of dementia and Alzheimer disease. Archives of Neurology, 62(10), 1556-1560; 
[3] Whitmer, R. A., Gunderson, E. P., Quesenberry, C. P., Zhou, J., \& Yaffe, K. (2007). Obesity in middle age and future risk of dementia: a 27 year longitudinal population based study. BMJ, 335(7630), 705. \\
\hline\hline
\end{tabular}
\caption{Examples of generated medical references using parametric citation methods. The citations are often incorrect and ill-formatted. Most importantly, they are hard to evaluate due to lack of API access with sources.}
\label{table:generated-citation-example}
\end{table*}





\section{Annotation Guidelines and Analysis}
\label{appendix:annotataion-guide}
Below we provide the guidelines we used for the human annotation in Section~\ref{para:human_eval}. We ask annotators to follow these guidelines to make an attribution judge.
\subsection{Annotation Guidelines}
\textbf{Citation Recall} measures how well the combination of all citations supports the statement.
\begin{itemize}
    \item For each statement, review all the provided citations (e.g., PubMed articles) as a group.
    \item Determine if the combined information from these citations fully supports, or cannot fully support the statement.
\end{itemize}

\textbf{Citation Precision} measures how well each individual citation supports the statement.
\begin{itemize}
    \item For each citation, evaluate whether it alone fully supports, partially supports, or does not support the statement.
    \item Repeat this evaluation for each citation independently.
\end{itemize}

\textbf{Note: Please only use the abstract of the PubMed article as a citation, not the whole body (only review the abstract instead of the whole article).}

\textbf{Clarification on "Fully Supported":} The determination depends on the relationship between the statement and the content in the citation(s).
\begin{itemize}
    \item \textbf{Words Not Mentioned in Articles:} If the word(s) in the statement represent something entirely different from what the article describes (e.g., distinct medical terms with no overlap), the statement cannot be considered "fully supported." In such cases, the support would likely be "not supported" or "partially supported", depending on how closely related the information is. 
    
    If the word(s) describe a subclass or specific instance of a broader concept mentioned in the article (e.g., the article discusses a class of treatments, and the statement mentions one treatment within that class), the citation may qualify as "partially supported".

    \item \textbf{Fully Supported Criteria:} A statement can only be considered "fully supported" if all key terms and concepts in the statement are directly addressed and explicitly supported by the information in the citation(s).
\end{itemize}

\subsection{Examples}
\begin{itemize}
    \item \textbf{Statement:} "Fruits like apples are a rich source of Vitamin C."
    \item \textbf{Citation 1:} The article mentions that "fruits like oranges, strawberries, and kiwis are excellent sources of Vitamin C".
    \item \textbf{Citation 2:} The article discusses "apples being nutritious but focuses on their fiber content", without mentioning Vitamin C.
\end{itemize}

\textbf{Recall (Combination of Citations):} If you look at both citations together, they do not fully support the statement. Although Citation 1 mentions fruits rich in Vitamin C, it does not explicitly include apples, and Citation 2 does not provide relevant information about Vitamin C in apples.

\textbf{Recall Score:} Not supported.

\textbf{Precision (Each Citation Individually):}
\begin{itemize}
    \item \textbf{Citation 1:} Provides "partial support" because it mentions fruits rich in Vitamin C but does not specify apples.
    \item \textbf{Citation 2:} Provides "no support" because it does not mention Vitamin C at all.
\end{itemize}

\section{Case Study: MedCite v.s. Open-domain Baselines}
\label{appendix:case-study}

To illustrate the limitations of existing baselines and the benefits of MedCite, we present a case study on the question in the BioASQ dataset: \textit{“Is avelumab effective for urothelial carcinoma?”}

We compare four configurations desccribed in Section \ref{sec:main-results}: \textbf{MedRAG}, \textbf{PRG}, \textbf{PGC}, \textbf{MedCite} with Llama-3-8B-I. as the backbone LLM.

Below we present excerpts from each system's output, followed by statement-level citation evaluation. Full outputs and supporting documents are provided in Appendix~\ref{appendix:case-study-full}.
\subsection{Analysis}

\begin{table*}[!ht]
\captionsetup{justification=raggedright,singlelinecheck=false}
\centering
\renewcommand{\arraystretch}{1.5}
\small
\begin{tabular}{
    m{2cm} 
    m{7cm} 
    m{6cm}
}
\hline\hline
\multicolumn{3}{c}{\textbf{Question: Is avelumab effective for bladder cancer?}} \\
\hline
\multicolumn{1}{c}{\textbf{Method}} & 
\multicolumn{1}{c}{\textbf{Generated Answer}} & 
\multicolumn{1}{c}{\textbf{Statement Evaluation}} \\
\hline
MedRAG & 
\textit{Avelumab is a PD-L1 inhibitor that has been approved by the FDA for the treatment of urothelial carcinoma (UC) in patients who have received platinum-based chemotherapy. The JAVELIN Bladder 100 trial showed that avelumab maintenance treatment after first-line chemotherapy significantly prolonged overall survival...} &
\begin{itemize}
    \item Statement 1: No citation — unverifiable
    \item Statement 2: Not supported — document coverage unclear
\end{itemize} \\
\hline
PRG & 
\textit{Avelumab is a PD-L1 inhibitor that has been shown to be effective in treating advanced urothelial carcinoma. In the JAVELIN Bladder 100 trial, avelumab was found to significantly prolong overall survival compared to best supportive care alone [17]. Additionally, avelumab has been approved by the FDA for the treatment of patients with locally advanced or metastatic urothelial carcinoma who have disease progression during or following platinum-containing chemotherapy [17].} &
\begin{itemize}
    \item Statement 1: Not Fully Supported — no citation provided
    \item Statement 2: Fully Supported
    \item Statement 3: Not Fully Supported — cited doc does not mention FDA approval
\end{itemize} \\
\hline
PGC & 
\textit{Avelumab is a monoclonal antibody that targets PD-L1, a protein that helps cancer cells evade the immune system [1][2]. It has been studied in various clinical trials for the treatment of bladder cancer, particularly in patients with locally advanced or metastatic urothelial carcinoma [3][4]. The results of these trials have shown that avelumab can improve overall survival and response rates [5][6]. Therefore, based on the available evidence, I conclude that avelumab is effective for bladder cancer [7][8].} &
\begin{itemize}
    \item All statements Fully Supported
\end{itemize} \\
\hline
MedCite & 
\textit{Avelumab is a PD-L1 inhibitor that has been shown to be effective in treating advanced urothelial carcinoma [2][3]. In the JAVELIN Bladder 100 trial, avelumab was found to significantly prolong overall survival compared to best supportive care alone [1][4]. Additionally, avelumab has been approved by the FDA for the treatment of patients with locally advanced or metastatic urothelial carcinoma who have disease progression during or following platinum-containing chemotherapy [1][5][6].} &
\begin{itemize}
    \item All statements Fully Supported
\end{itemize} \\
\hline\hline
\end{tabular}
\caption{Comparison of model-generated answers and statement evaluation for an example.}
\label{table:case-study}
\end{table*}

This case highlights several key limitations of existing methods and the advantages of MedCite. MedRAG and PRG both rely on a single round of question-level retrieval. As a result, important claims may be missing supporting documents entirely (MedRAG), or incorrectly cited due to coarse retrieval granularity (PRG). In contrast, PGC performs better in terms of support coverage: all statements are supported through post-hoc citation retrieval. However, its generation is purely based on parametric model knowledge, making it difficult to update and prone to factual drift as medical knowledge evolves.

MedCite bridges these gaps. By grounding the initial answer generation in retrieved documents (like PRG), and then refining citation support through statement-level retrieval and reranking (like PGC), it achieves the best of both worlds: answer correctness, completeness, and adaptability. In high-stakes domains like medicine, where knowledge updates are frequent and trustworthiness is paramount, this hybrid design offers a more robust and verifiable solution than purely parametric or coarse-retrieval-based approaches.

\subsection{Case Study Full Outputs and Citations}
\label{appendix:case-study-full}
\begingroup
\renewcommand{\arraystretch}{1.0}
\begin{table*}[!ht]
\centering
\small

\begin{tabular}{p{0.975\textwidth}}
\hline\hline
\textbf{Question:} Is avelumab effective for bladder cancer? \\
\hline
\textbf{MedRAG Answer:} \\
After reviewing the documents, I found that avelumab is a PD-L1 inhibitor that has been approved by the FDA for the treatment of urothelial carcinoma (UC) in patients who have received platinum-based chemotherapy. The JAVELIN Bladder 100 trial showed that avelumab maintenance treatment after first-line chemotherapy significantly prolonged overall survival compared to best supportive care alone. I also found that avelumab has been shown to be effective in treating patients with locally advanced or metastatic UC who have responded or stabilized after first-line platinum-containing chemotherapy. The National Comprehensive Cancer Network Clinical Practice Guidelines in Oncology recommend avelumab as a category 1 preferred checkpoint inhibitor maintenance option for patients with cisplatin-eligible and cisplatin-ineligible locally advanced or metastatic UC. Based on the evidence, I conclude that avelumab is effective for bladder cancer. \\
\hline
\textbf{PRG Answer:} \\
Avelumab is a PD-L1 inhibitor that has been shown to be effective in treating advanced urothelial carcinoma. In the JAVELIN Bladder 100 trial, avelumab was found to significantly prolong overall survival compared to best supportive care alone [17]. Additionally, avelumab has been approved by the FDA for the treatment of patients with locally advanced or metastatic urothelial carcinoma who have disease progression during or following platinum-containing chemotherapy, or have disease progression within 12 months of neoadjuvant or adjuvant platinum-containing chemotherapy [17]. \\
\textbf{Cited Documents}:\\
Document [17] (Title: Avelumab Maintenance Therapy for Advanced or Metastatic Urothelial Carcinoma, PMID: 32945632): "Platinum-based chemotherapy is standard-of-care first-line treatment for advanced urothelial carcinoma. However, progression-free survival and overall survival are limited by chemotherapy resistance. ..." \\
\hline
\textbf{PGC Answer:} \\
Avelumab is a monoclonal antibody that targets PD-L1, a protein that helps cancer cells evade the immune system [1][2]. It has been studied in various clinical trials for the treatment of bladder cancer, particularly in patients with locally advanced or metastatic urothelial carcinoma [3][4]. The results of these trials have shown that avelumab can improve overall survival and response rates [5][6]. Therefore, based on the available evidence, I conclude that avelumab is effective for bladder cancer [7][8]. \\
\textbf{Cited Documents}:\\
Document [1] (Title: Immune checkpoint inhibitors of PD-L1 as cancer therapeutics, PMID: 31488176): "Since the discovery of immune checkpoint proteins, there has been a special interest in developing antibodies that block programmed cell death 1 receptor (PD-1) and programmed cell death receptor ligand 1 (PD-L1) ..."\\
Document [2] (Title: Avelumab: combining immune checkpoint inhibition and antibody-dependent cytotoxicity, PMID: 28274143): "Immune checkpoint inhibition holds great promise for selected tumors. ..."\\
Document [3] (Title: Immunotherapy in bladder cancer-quo vadis? Update on current trials and developments, PMID: 32468092): "Great advances have been made for the treatment of urothelial carcinoma by the introduction of checkpoint inhibitors (CPI). ..."\\
Document [4] (Title: The biology and rationale of targeting nectin-4 in urothelial carcinoma, PMID: 33239713): "Bladder cancer is the tenth most common cancer type worldwide. Urothelial carcinoma is the most common type of bladder cancer and accounts for 90\% of bladder cancer cases in the USA and Europe. ..."\\
Document [5] (Title: Avelumab for advanced Merkel cell carcinoma in the Netherlands: a real-world cohort, PMID: 32948651): "Merkel cell carcinoma (MCC) is associated with high recurrence rates and poor survival when metastatic disease is present. ..."\\
Document [6] (Title: Clinical Evaluation of Avelumab in the Treatment of Advanced Urothelial Carcinoma: Focus on Patient Selection and Outcomes, PMID: 35237074): "First-line therapy for treatment of advanced urothelial carcinoma includes combination platinum-based chemotherapies, ..."\\
Document [7] (Title: Plain language summary of results from the JAVELIN Bladder 100 study: avelumab maintenance treatment for advanced urothelial cancer, PMID: 35416053): "This is a plain language summary of an article originally published in <iThe New England Journal of Medicine</i. ..."\\
Document [8] (Title: Cost-Effectiveness of Avelumab Maintenance Therapy Plus Best Supportive Care vs. Best Supportive Care Alone for Advanced or Metastatic Urothelial Carcinoma, PMID: 35570929): "Avelumab (MSB0010718C) is a fully human anti-programmed cell death ligand 1(PD-L1) antibody against PD-L1 interactions and enhances immune activation against tumor cells in the meantime. ..."\\
\hline\hline
\caption{Case study answers and cited documents.}
\end{tabular}

\label{tab:case-study-first-three}
\end{table*}
\endgroup

\newpage
\begingroup
\renewcommand{\arraystretch}{1.0}
\begin{table}[!ht]
\centering
\small
\begin{tabular}{p{0.975\textwidth}}
\hline\hline
\textbf{MedCite:} \\
Avelumab is a PD-L1 inhibitor that has been shown to be effective in treating advanced urothelial carcinoma [2][3]. In the JAVELIN Bladder 100 trial, avelumab was found to significantly prolong overall survival compared to best supportive care alone [1][4]. Additionally, avelumab has been approved by the FDA for the treatment of patients with locally advanced or metastatic urothelial carcinoma who have disease progression during or following platinum-containing chemotherapy, or have disease progression within 12 months of neoadjuvant or adjuvant platinum-containing chemotherapy [1][5][6]. \\
\textbf{Cited Documents}:\\
Document [1] (Title: Avelumab Maintenance Therapy for Advanced or Metastatic Urothelial Carcinoma, PMID: 32945632): "Platinum-based chemotherapy is standard-of-care first-line treatment for advanced urothelial carcinoma. ..." \\
Document [2] (Title: Clinical Evaluation of Avelumab in the Treatment of Advanced Urothelial Carcinoma: Focus on Patient Selection and Outcomes, PMID: 35237074): First-line therapy for treatment of advanced urothelial carcinoma includes combination platinum-based chemotherapies, though resistance and long-term toxicity concerns to these regimens cause limitations in progression-free survival and overall survival. ..." \\
Document [3] (Title: Which place for avelumab in the management of urothelial carcinoma?, PMID: 31286802): "<bIntroduction</b: Urothelial carcinoma (UC) has a poor prognosis, with the only standard first-line metastatic treatment being platinum-based chemotherapy. ..." \\
Document [4] (Title: Patient-reported Outcomes from JAVELIN Bladder 100: Avelumab First-line Maintenance Plus Best Supportive Care Versus Best Supportive Care Alone for Advanced Urothelial Carcinoma, PMID: 35654659): In JAVELIN Bladder 100, avelumab first-line maintenance plus best supportive care (BSC) significantly prolonged overall survival (OS; primary endpoint) versus BSC alone in patients with advanced urothelial carcinoma (aUC) without disease progression with first-line platinum-containing chemotherapy. ..." \\
Document [5] (Title: FDA Approval Summary: Atezolizumab for the Treatment of Patients with Progressive Advanced Urothelial Carcinoma after Platinum-Containing Chemotherapy, PMID: 28424325): "Until recently in the United States, no products were approved for second-line treatment of advanced urothelial carcinoma. ..." \\
Document [6] (Title: Avelumab in metastatic urothelial carcinoma after platinum failure (JAVELIN Solid Tumor): pooled results from two expansion cohorts of an open-label, phase 1 trial, PMID: 29217288): "The approval of anti-programmed death ligand 1 (PD-L1) and anti-programmed death 1 agents has expanded treatment options for patients with locally advanced or metastatic urothelial carcinoma. ..." \\
\hline\hline
\caption{Case study answers and cited documents.}
\end{tabular}

\end{table}
\endgroup

\end{document}